\documentclass[onecolumn]{IEEEtran}

\usepackage{comment}
\usepackage{mathtools}
\usepackage{amsmath}
\usepackage{multirow}
\usepackage{multicol}
\usepackage{booktabs}
\usepackage{graphicx}
\usepackage{subcaption}
\usepackage[ ] {algorithm2e}
\usepackage{tikz}
\usepackage{textcmds}
\usepackage{mhchem}
\usepackage{dirtytalk}
\usepackage{csquotes}
\usepackage{float}

\usepackage{array}
\usepackage{units}
\usepackage{multirow}
\usepackage{babel}
\makeatother
\usepackage{amssymb}
\usepackage{babel}
\usepackage{bbding}

\ifCLASSOPTIONcompsoc
  \usepackage[nocompress]{cite}
\else
  \usepackage{cite}
\fi

\ifCLASSINFOpdf

\else

\fi

\begin{document}


\title{Spike-based Neuromorphic Computing for  Next-Generation Computer Vision}

\author{
\IEEEauthorblockN{
Md Sakib Hasan, Catherine D. Schuman, Zhongyang Zhang, Tauhidur Rahman, and Garrett S. Rose
}

}



\IEEEtitleabstractindextext{%
\begin{abstract}
Neuromorphic computing promises orders of magnitude improvement in energy efficiency compared to the traditional von Neumann computing paradigm. The goal is to develop an adaptive, fault-tolerant, low-footprint, fast, low-energy intelligent system by learning and emulating brain functionality, which can be realized through innovation in different abstraction layers, including material, device, circuit, architecture, and algorithm. As the energy consumption in complex vision tasks keep increasing exponentially due to larger data sets and resource-constrained edge devices become increasingly ubiquitous, spike-based neuromorphic computing approaches can be viable alternatives to deep convolutional neural networks that are dominating the vision field today. In this chapter, we introduce  neuromorphic computing, outline a few representative examples from different layers of the design stack (devices, circuits, and algorithms) and conclude with a few exciting applications and future research directions that seem promising for computer vision in the near future. 
\end{abstract}

}

\maketitle
\IEEEdisplaynontitleabstractindextext
\IEEEpeerreviewmaketitle

\section{Introduction}
\label{intro}
Since Gordon Moore's famous prediction in 1965 \cite{moore2006cramming}, colloiquially known as Moore's Law, the computing industry has worked relentlessly to achieve exponential increase in speed and functional density over last six decades using progressively advanced fabrication technologies along with architectural innovations. Moore's law along with Dennard scaling \cite{dennard1974design} resulted in the doubling of performance per joule about every 18 months. Unfortunately, Dennard scaling started breaking down in the early 2000s, resulting in a saturation of single core frequency due to the power constraint popularly known as the power wall \cite{patterson2016computer}.  Another major bottleneck of current digital computers is the separation of memory and processor in the conventional von Neumann architecture. The lagging behind of memory speed compared to processors has given rise to von Neumann bottleneck also known as the memory wall, that is, most energy and time is used in trafficking data between these two subsystems instead of the actual computation \cite{patterson2016computer}. Moreover, the scaling of transistor feature size as dictated by Moore's law has also been slowing down and may reach an end in the coming decade due to physical limits \cite{williams2017s}.

With the ever-increasing demand of computing in this data-intensive world looking forward to a future with billions of smart interconnected devices known as Internet-of-Things (IoT) along with the advent of a new age of artificial intelligence (AI), researchers are looking into novel solutions for sustaining the computing revolution. Diverse approaches, such as optical \cite{solli2015analog}, quantum \cite{preskill2018quantum}, biomolecular \cite{puaun1998dna}, and neuromorphic \cite{mead1990neuromorphic} have been explored to go beyond the traditional paradigm. In this chapter, we focus on spike-based neuromorphic computing, which is a bio-inpsired approach 
that tries to emulate excellent energy efficiency and superior information processing capabilities that can be observed in biological organisms\cite{herculano2016human}.

The early work in modeling biological neural networks with electrical circuits can be traced back to pioneering work by McCulloch and Pitts in \cite{mcculloch1943logical}. Another seminal work was  Hebbian learning \cite{Hebb1949TheTheory} in 1949, which is a classic inspiration for spiking neural network (SNN) researchers. 1952 saw a big breakthrough with the development of Hodgkin-Huxley neuron models \cite{hodgkin1952quantitative}, which shed light into the complex dynamics of this biological computational primitive followed by subsequent models \cite{fitzhugh1961impulses,nagumo1962active}. Though not the predominant research direction in the AI community (being overshadowed by symbolic AI and other directions), a few notable AI researchers began exploring neural networks as a possible route towards building an intelligent machine which led to seminal works such as perceptron \cite{Rosenblatt1958}, multilayer networks  \cite{werbos1974beyond}, backpropagation training \cite{rumelhart1986learning}, the Hopfield network \cite{hopfield1982neural} and self-organizing maps \cite{kohonen1982self}, among many others.

However, the dream of neuromorphic computing is to combine brain-inspired algorithm and hardware together to build a different class of intelligent system. This concept was first proposed by Carver Mead at Caltech in the late 1980s \cite{mead2020we}. As a pioneer in the VLSI (very large sale integration) digital computer revolution, Mead realized some of its limitations and started exploring an analog/
mixed-signal design paradigm to emulate biological functions with the electronics of an integrated circuit (IC)  \cite{mead1990neuromorphic}. Since its inception, vision has been a vibrant field of study in neuromorphic computing as can be seen in the celebrated early works in building the silicon retina in 1994 \cite{mahowald1994silicon} which followed the famous silcion neuron from 1991\cite{mahowald1991silicon}. More recently, the world has seen large-scale neuromorphic processors such as BrainScaleS \cite{schemmel2010wafer}, TrueNorth \cite{merolla2014million}, Neurogrid \cite{benjamin2014neurogrid}, SpiNNaker \cite{furber2014spinnaker}, and Loihi \cite{davies2018loihi}. The research in this field has primarily two thrusts: (1) learning the working principle behind human perception and cognition, a fundamental scientific question of perennial interest, specially to cognitive neuroscientists, (2) building a new class of computing machines overcoming the limitations of traditional von Neumann digital computers, which is of primary interest to researchers working in the frontier of computing technology. Since building such machine requires understanding and innovation across the entire design hierarchy, namely, materials, device, circuit, system, architecture, communication, and algorithm along with a deep understanding of neuroscience principles, neurmorphic computing has become a vibrant interdisciplinary research endeavor over the last three decades. 

The early work on neuromorphic computing \cite{mead1989analog} explored deep similarity between conduction in electronics and  ion-channel dynamics of biological neural networks based on the physics of electronic components such as transistors under special operating conditions. The term has gradually evolved to describe a set of brain-inspired hardware and algorithms for neural networks with varying degrees of biofidelity. Modern digital computers usually store information using 32 or 64 bits and process it using synchronous deterministic architecture with separate memory and processing units made of solid-state electronic devices. On the other hand, our brains use patterns of neuron spikes to represent and process information using stochastic computational elements made of organic materials with collocated memory and processing units. In the literature, designs lying anywhere between traditional digital computer and brain-like architecture have been termed as neuromorphic computing. Since we cannot do justice to this wide variety of works, we focus on SNNs for
vision applications in this chapter for the sake of brevity and coherence.

The rest of this chapter is organized as follows: Section \ref{sec_back} gives basic background on  biological neural networks  neuromorphic computing systems. Section \ref{sec_dev} discusses several traditional and emerging devices and circuits with significant promise for building scalable neuromorphic systems with enhanced functionality followed by a brief summary of state-of-the-art neuromorphic processors in Section \ref{sec_proc}.  Section \ref{sec_alg} discusses several architecture and algorithms using SNNs, which are the principal abstraction
of the nervous system that neuromorphic computing systems employ to emulate brain function.
In Section \ref{sec_app}, we discuss several promising vision applications including ongoing research on dancing pose estimation. Finally, Section \ref{sec_con} concludes this chapter with a summary along with directions for future research.

\section{Background on Neuromorphic Computing and Bio-inspired Spiking Neural network }
\label{sec_back}

In this section, we will first outline the main characteristics of a neuromorphic system followed by a concise overview of the working principle of biological neural network for inspiration. Then we briefly outline the defining aspects of the spiking neuromorphic computing system to distinguish it from traditional digital computer as well as the artificial neural network (ANN) using neurons with a continuous activation function without temporal dynamics.   

\subsection{Characteristics of Neuromorphic Computing System}

There are several brain-inspired characteristics that distinguish a neuromorphic computer from a traditional computer. First, it should have a brain-like, massively parallel operation unlike the traditional von Neumann single core CPU, which was developed using a \lq stored program\rq\  model for strictly sequential computation. Second, it will have collocated memory and processing, aka CIM (computation in memory) analogous to synapses and neurons in a biological brain overcoming the \lq memory wall\rq\  resulting from the von Neumann architecture with separate units for storage and computation. For example, synaptic weight serves both as a memory and a computational element where presynaptic spikes produce currents that lead to an increase in the post-synaptic membrane potential. Third, asynchronous and/or analog/mixed-signal computation as opposed to globally synchronized digital computation in traditional platforms. In brain, synapses can store and perform analog operation on the input data; neurons produce binary spikes but they have internal temporal states which are analog along with information encoded in time interval between spikes which is also analog; no global clock. Any one of the four possible combinations between synchronous/asynchronous and analog/digital is possible with its unique trade-offs and different combinations have been used for different neuromorphic processors.  Fourth, sparsity in connection and activation, which is essential for energy-efficient computation and large-scale network like brain. Additionally, stochasticity and nonlinear dynamics play a key role in the operation of biological brain, and the learning mechanism is almost certainly not global and mostly unsupervised unlike supervised global algorithm such as backpropagation widely used in ANN. This is an active field of research and a thorough understanding still eludes the research community. Of course, there are other features but achieving these above-mentioned properties through algorithm and hardware innovation remains the primary thrust.   

\subsection{Biological Underpinning}

\begin{figure}
\centering
  \subcaptionbox{}{\includegraphics[width=4.5 in]{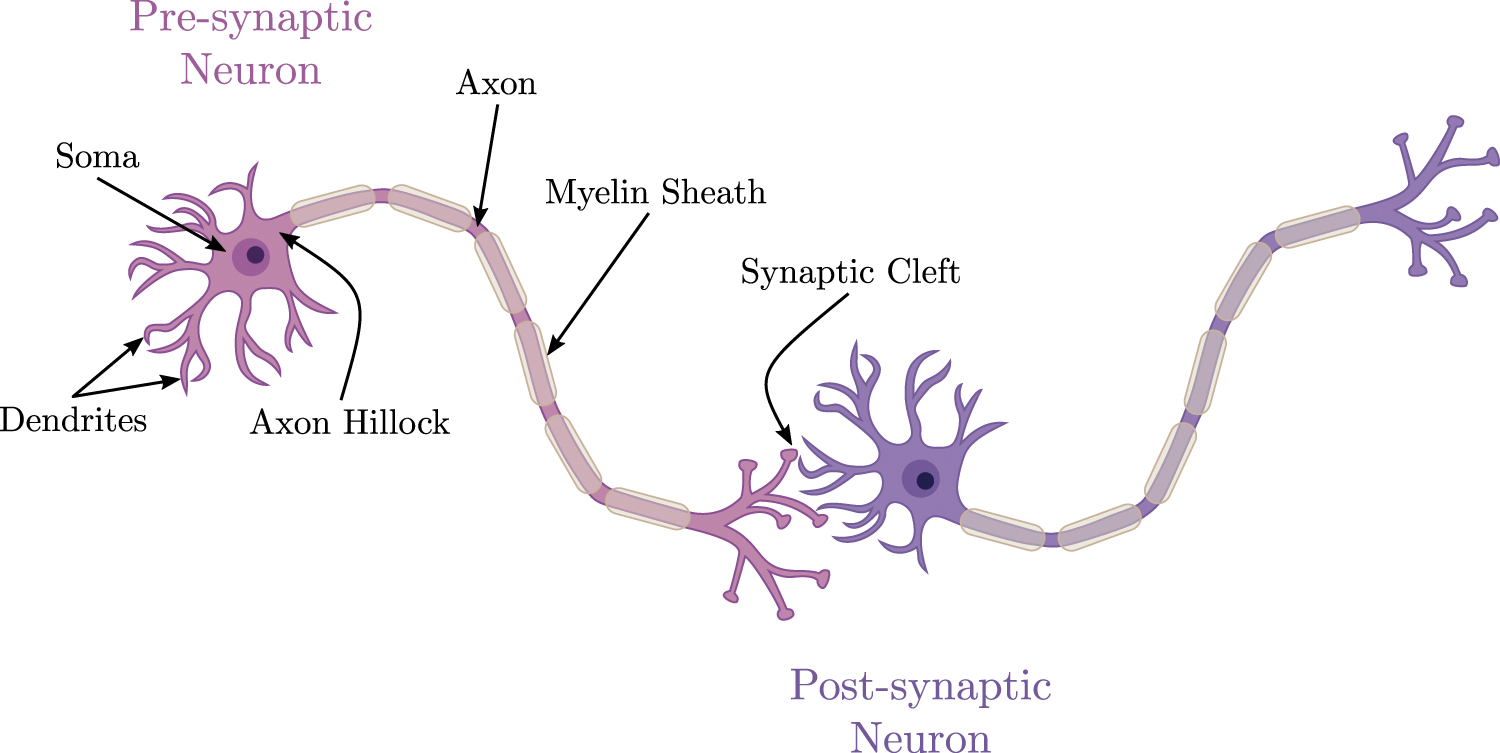}}%
  \label{1_a}
\hspace{-2.5em}%
  \subcaptionbox{}{\includegraphics[width=2.25 in]{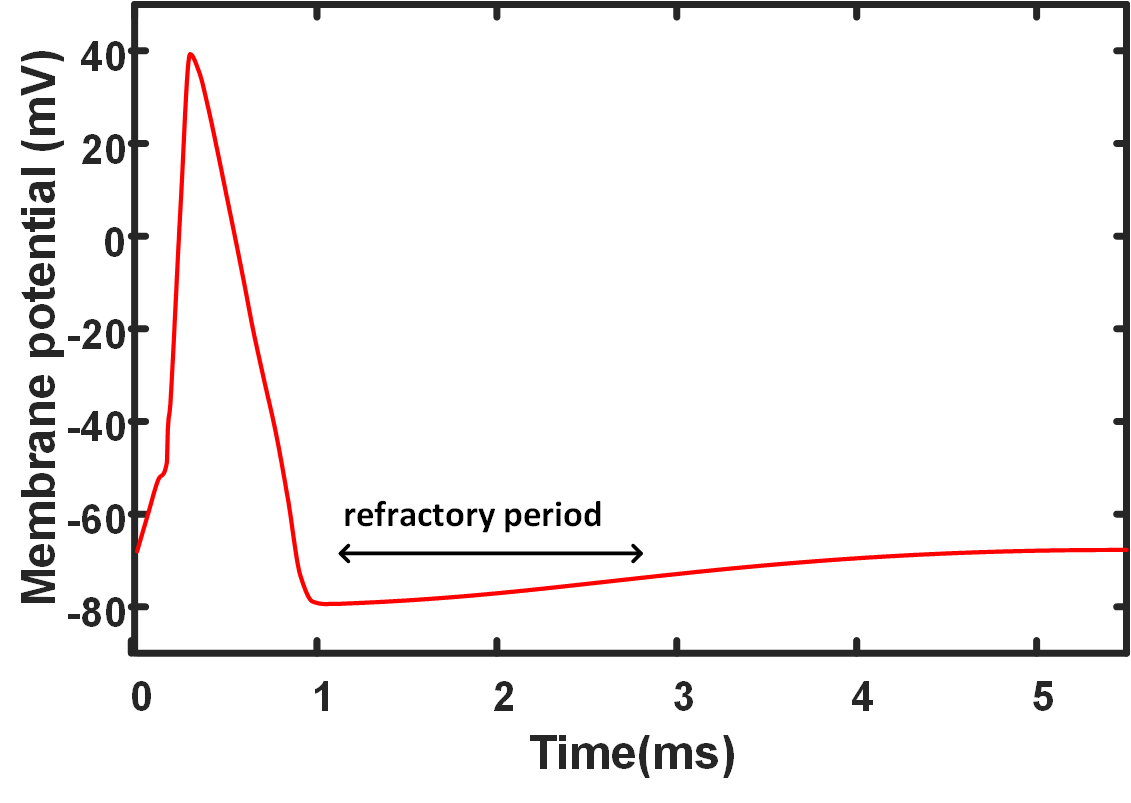}}%
  \subcaptionbox{}{\includegraphics[width=2.25 in]{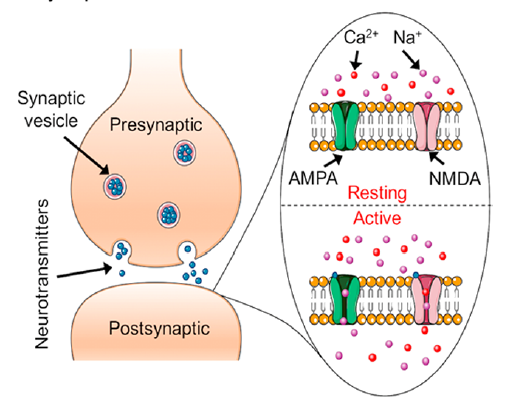}}%
  \caption{(a) Simplified neural communication schematic \cite{rose2021system} (Reproduced from Rose et al., Neuromorphic Computing and Engineering 1.2 (2021). Copyright CC BY 4.0.), (b) Post-synaptic spike generation, (c) Chemical synapse \cite{Najem2018MemristiveMimics}; reproduced with permission from Najem et al., ACS nano 12.5 (2018), Copyright 2018
American Chemical Society.}
  \label{fig_1}
\end{figure}

The generation and transmission of action potentials in biological neural networks is a complex dynamical process. Here, we present a significantly simplified account to convey the big picture. Neurons are tiny cells that respond to spiking electrochemical stimuli by generating their own spiking outputs.  Biological neural networks are networks of neuron. Figure \ref{fig_1}(a) shows two neurons connected to each other. The pink neuron (to the left) produces a spiking response that flows into the purple neuron (to the right) via synaptic cleft which is commonly known as the synapse \cite{rose2021system}. In this setup, the pink one is the pre-synaptic neuron and the purple one is the post-synaptic neuron. Synapses are responsible for regulating the impact an incoming signal will have on the post-synaptic neuron. This regulation is often modeled as a weight that either allows spiking signals to pass through virtually unobstructed or provides some resistance that essentially dulls the signal strength. Much of our ability to learn has been attributed to this synaptic regulation, as well as the creation or removal of synaptic connections.

A prototypical spiking neuron has four main parts: dendrites, soma, axon, and synapse. Dendrites together with the soma act as a leaky integrator of ionic currents that flow into or out of the neuron. These currents result from the opening and closing of channels in the cell membrane and through ion diffusion or active ion pumps. For example, when a presynaptic neuron creates an action potential, neurotransmitters are released and cause the opening of ligand-gated ion channels at the post-synaptic neuron’s dendrites. This causes ions to flow resulting in an increase (in the case of an excitatory pre-synaptic neuron) or a decrease (in the case of an inhibitory presynaptic neuron) in the potential of the post-synaptic neuron’s cell membrane. Interestingly, a number of complex computations such as boolean logic functions can take place in the dendritic arbor before information reaches the cell body. If the membrane potential rises (also called depolarization) from its resting value of around -70 mV to a threshold value (typically around -55 mV), then an action potential about 1 ms wide and +40 mV at its peak as shown in Figure \ref{fig_1}(b) is generated, which travels down the neuron’s axon, ending at the synaptic terminals where neurotransmitter is released to the next set of neurons. The length of an axon makes it analogous to a transmission line that will provide some delay and attenuation of signals as they propagate from one neuron to another. After a neuron spikes, there is typically a refractory period of several milliseconds during which it cannot produce another spike. During this time, the neuron’s membrane is hyperpolarized below its resting potential.

Neurons communicate with each other using two types of synapses, namely, electrical and chemical synapses. Chemical synapses play a major role in both communication and learning in the central nervous system (Figure \ref{fig_1}(c)). Generation of action potential in a pre-synaptic neuron causes an influx of calcium ions followed by docking of synaptic vesicles at the cell membrane and releases neurotransmitters into the synaptic cleft. We would like to mention two neurotransmitters, namely, glutamate and $\gamma$-aminobutyric acid (GABA), which exert excitatory and inhibitory effect on the post-synaptic membrane potential, respectively. The released neurotransmitters are received at the post-synaptic plasma membrane by AMPA and NMDA receptors/ions channels, events that trigger ion flux into the post-synaptic neuron that depolarizes the cell \cite{krnjevic1974chemical}. Importantly, these channels remain closed, holding the membrane in an insulating state, until NTs bind to their receptors, at which point the conductance of the post-synaptic
membrane increases exponentially. Other additional factors influencing neural dynamics included relative timing of pre- and post-synaptic neuron activity, which can result in long-term strengthening  or, weakening of synaptic connections known as LTP (long-term potentiation) and LTD (long-term depression), respectively. One such example is famous Hebbian learning rule popularly phrased as ``neurons that fire together, wire together" \cite{Hebb1949TheTheory}. Finally, electrical synapse also joins neurons via gap junction ion channels enabling fast, threshold-independent, and bidirectional ion transport  and this synaptic strength can be modulated significantly via activity-dependent plasticity with a significant impact on the synchronization in mammalian neural networks \cite{haas2011activity} \cite{haas2016activity}.

\subsection {Spiking Neural Network}
The seminal paper from Maass \cite{maass1997networks} divided neural netowrks into three generations based on the neuronal dynamics. The first generation is named McCulloch–Pitt perceptrons, based on  the simple McCulloch–Pitt thresholding neuron operation \cite{mcculloch1943logical}. The second generation neuron has continuous nonlinear activation functions such as tanh, ReLU etc. which enables it to evaluate a continuous set of output values, enables gradient descent based backpropagation due to its differentiable nature, and has propelled the  current boom in Artificial Neural Network (ANN) \cite{rumelhart1986learning}. The third generation of networks use spiking neurons primarily of the ‘integrate-and-fire’ (IF) type \cite{izhikevich2003simple}, which communicate using spikes and have the promise the create the next paradigm shift in AI.

Biological intelligence emerges from the computation carried out by networks of neurons communicating through spikes. State-of-the-art ANNs significantly
abstract the behavior of biological neuronal networks, with several simplifications such as reducing a neuron’s behavior to a simple spike rate. However, the performance of ANNs in the field of computer vision tasks is less efficient than their biological counterparts in terms of power and speed \cite{frenkel2021bottom}. Biological brain processes information in a massively parallel asynchronous manner, whereas deep neural networks (DNN), even in parallel multi-core computing platforms, compute in a essentially sequential form since each layer's computation has to be completed before starting the computation in the next layer and can only parallelize operations in the same layer. The situation is much worse in traditional single core system. As a result, ANN can suffer from significant delay compared to SNN \cite{camunas2019neuromorphic}. In 1996, Thorpe et al.\cite{thorpe1996speed} showed that the biological brain is able to recognize visual images with one spike propagating through all layers of the visual cortex, and Rolls and Tovee \cite{rolls1994processing} measured the same visual processing speed in the macaque
monkey. Theses works highlight the amazing efficiency of the spike-based information encoding technique in brains.

The efficiency of spike-based computation motivated machine learning (ML) and neuroscience researchers to
begin exploring SNNs, the third generation of neural networks. Computation in SNNs is event-driven as in the biological brain, so each neuron in the network generates its outputs only when enough spikes indicating the existence of a specific feature or pattern have been detected \cite{camunas2019neuromorphic}. This feature gives SNNs the capability to solve complex spatiotemporal tasks and to make use of efficient event-driven sensors, such as event-based cameras.

\begin{table}
  \centering
  \caption{Comparison between ANN and SNN}
\scalebox{1.2}{
  \renewcommand{\arraystretch}{1.3}
  
\begin{tabular}{|c|c|c|}
\hline 
Feature & ANN & SNN \tabularnewline
\hline 

Data & Static Frame-based & Dynamic Event-based\tabularnewline
\hline 

Neuronal activation & Continuous-valued real number & Discrete-valued spike \tabularnewline
\hline 

Differentiable & Yes & No \tabularnewline
\hline 

Short-term memory & Network & Synapse, neuron and network \tabularnewline
\hline

Computational Complexity & Moderate & High (greatly benefited from compute-in-physics) \tabularnewline
\hline

Biofidelity & Low & Moderate/ high  \tabularnewline
\hline

\end{tabular}}
\label{tab:SNN_DNN}
\end{table}

Table \ref{tab:SNN_DNN} shows a comparison of the main aspects between some of the key properties of ANNs and SNNs for vision \cite{hendy2022review}. As stated in the previous section, synchronous computation in each layer of DNNs can be time consuming. On the other hand, in SNNs, the computation is processed asynchronously in spike form, allowing information to propagate to the next layer before all computation in the current layer is complete. However, this asynchrony combined with the nondifferentiable nature of spikes complicates the credit assignment problem and limits the use of many popular training algorithms employed in DNNs. On the other hand, the inherent temporal dynamics of SNNs allows them to perform more complex tasks than DNNs \cite{maass1997networks}. For
example, SNNs have neuron-level temporal memory, enabled by the leaky integration of information at the neuron’s input. This means that even purely feed-forward SNNs have an inherent short-term memory. Contrast this with DNNs, which can have short-term memory enabled by their network topology (e.g., with recurrent connections), but there is usually no built-in temporal memory at the level of individual neurons. This extra layer of short-term memory in SNNs makes them a good fit for temporal processing of data such as audio and video.

A big challenge in the area of neuromorphic computing is determining how much detail of the physiological processes needs to be modeled to faithfully capture the underlying computational principles.  Popular ANN neuron models such as such as tanh and ReLUs lose temporal information and only convey information about the spike rate. Due to their computationally efficiency and differentiable nature, they are employed in most modern DNNs. The spiking neuron models used in neuromorphic computing need to have enough complexity to capture key dynamic properties of biological neurons without having significant computational or hardware overhead. In general, adding more biological features leads to exponential growth in the computational cost. Complex neuron models such as the multi-compartment Hodgkin–Huxley model \cite{hodgkin1952quantitative}, and the FitzHugh–Nagumo model \cite{fitzhugh1961impulses}, \cite{nagumo1962active}  capture several complex dynamics behavior of spiking neurons. Unfortunately, their exceeding computational cost renders them mostly suitable for research in neuroscience. On the other hand, very simple threshold models such as the McCulloch–Pitts \cite{mcculloch1943logical} capture only simple neuron behavior, such as spiking above or below a particular rate.  It should be noted that calculating implementation cost in the traditional digital computer is not the only way forward. As we will see in Section \ref{sec_mem}, to bypass the digital computational cost incurred by complex neuron models, researchers are exploring analog nanoscale low-power scalable emerging neuromoprhic devices such as memristors, which can natively perform some of these complex neuronal dynamics via intrinsic device physics (known as compute-in-physics), which can reduce these costs by orders of magnitude.

\begin{figure}
    \centering
    \includegraphics[width=0.64\textwidth]{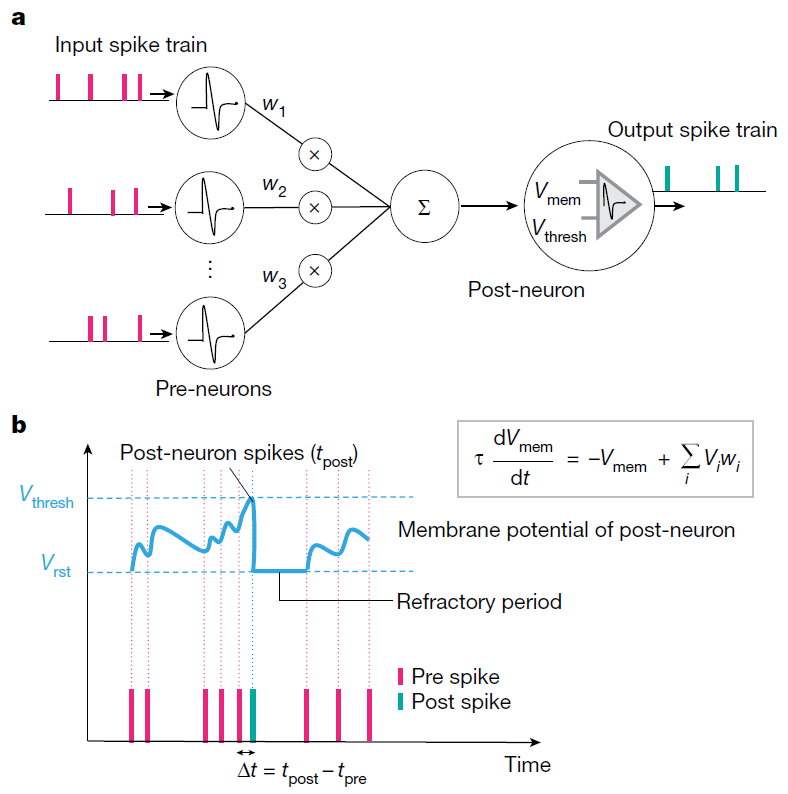}
    \caption{SNN with LIF (Leaky integrate and fire) neuron, (a) construction and (b) dynamics.  \cite{roy2019towards}. Reproduced with permission from Roy et al., Nature 575.7784 (2019). Copyright 2019 Springer Nature.}
    \label{fig:LIF}
\end{figure}

IF or leaky IF (LIF) is one of the most popular spiking neuron models used in neuromorphic computing since it exhibits adequate complexity to capture important temporal information of spike statistics, but abstracted enough to be computationally efficient and suitable for simple hardware implementation. Figure \ref{fig:LIF} illustrates a LIF-based SNN comprising a post-neuron driven by input pre-neurons. The pre-neuronal spikes, $V_i$, are modulated
by synaptic weights $w_i$ to produce a resultant current, $\sum_i V_i \times w_i$ (equivalent to a dot-product operation) at a given time. The resulting current affects the membrane potential of the post-neuron. Additionally, the dynamics of LIF spiking neurons are shown. The membrane potential, $V_{mem}$, integrates incoming spikes and leaks
with time constant, $\tau$ in the absence of spikes. The post-neuron generates an
outgoing spike whenever $V_{mem}$ crosses a threshold, $V_{thresh}$. A refractory period ensues after spike generation, during which $V_{mem}$ of the post-neuron is not affected. However, it is not able to capture some of the more complex biological features of real spiking
neurons such as phasic spiking, tonic and phasic bursting, spike frequency adaptation, and accommodation, to name a few. More detailed neuron models such as the Hodgkin–Huxley model \cite{hodgkin1952quantitative}, which requires numerical solution of four nonlinear differential equations, have significant computational overhead, making it difficult to employ them in low-power neuromorphic systems. Moreover, it is still unclear which biological features are necessary for designing neuromorphic systems with a particular set of desired behaviors. A number of other models have been proposed in the literature that have medium complexity and are a good tradeoff between feature richness and computational cost. These include models such as the adaptive IF model \cite{brette2005adaptive}, the spike response model \cite{gerstner1993spikes}, and the Izhikivich model \cite{izhikevich2003simple}, which are able to capture more complex dynamics than the LIF model, such as bursting and chattering. Still, other models capture additional behaviors that are important for
neural computation, such as stochasticity of spiking and synaptic transmission \cite{gerstner2002spiking} or energy dependence of neural activity \cite{burroni2017energetic}. An excellent review of these features and the associated neuron models can be found in \cite{izhikevich2004model}.

\section{Neuromorphic Devices and Circuits} 
\label{sec_dev}
SNN model complexity is significantly more than ANN due to dynamic behavior of building blocks such as synapses and neurons. Solving the constituent equations using digital electronics can lead to prohibitive computational cost nullifying the energy and information processing advantages of this emerging paradigm. Hence, analog circuit implementations of synapses and neurons capable of natively performing these computations can be particularly beneficial. Here, we first introduce traditional MOSFET based building blocks and then use memristor as a representative example of an emerging beyond-CMOS device with great promise for building next-generation neuromorphic computing systems.

\subsection{Synapses and Neurons using MOSFET}

\subsubsection{Synapse}
Several synaptic circuit using traditional CMOS circuits have been reported in the literature. Indiveri et al. reported synapses in 800 nm process technology with both  short and long-term
plasticity \cite{indiveri2006vlsi}. Another analog CMOS synapse was reported in \cite{koickal2007analog} with on-chip STDP (spike time dependent plasticity) learning. Similarly there have been several other works such as \cite{ebong2011cmos, tovar2008analog}, etc. In recent years, more attention has been focused towards NVM (Non-Volatile Memory) cross-bar architecture, which has great promise for efficient synaptic implementation for in-memory computing.

\subsubsection{Neuron}
Researchers have explored several traditional CMOS-based synapse and neuron circuits over the years since the early days of Mead and others \cite{mead1990neuromorphic,mead1989analog}. In 2003, Indiveri introduced an IF neuron with spike frequency adaptation and a configurable refractory period  with lower power consumption compared to existing axon hillock designs \cite{indiveri2003low}. Later, Indiveri et al. proposed a current-mode conductance-based IF silicon with plasticity for learning \cite{indiveri2010spike}. Another IF neuron operating in two asynchronous phases, integration phase followed by firing was reported in \cite{wu2014energy}. Since the inception of this field, subthreshold conduction in MOSFET has attracted a lot of attention since the low power exponential characteristics have certain similarity with ion-channel dynamics in biological neurons, and the reduced speed is a reasonable trade-off considering the desired biomimetic time scale. Ref. \cite{liu2014event} is a valuable resource for the interested reader willing to delve deeper into many designs resulting from this decades-long exploration. Besides, there have been several digital implementations reported in the literature trading cost and complex native dynamics for robust scalable operation \cite{skrbek1999fast, hikawa2003digital, muthuramalingam2008neural}. A mixed mode neuron  has also been reported \cite{sayyaparaju2020device} with on-chip tunability of accumulation rate enabling flexible interfacing with different types of devices.

\subsection{Synapses and Neurons Using Memristors}
\label{sec_mem}
During the last 15 years, quite a few emerging devices have been explored as potential candidates for neuromorphic computing. Emerging memory devices capable of non-volatile storage of analog values along with extreme density advantages are highly desirable to be used as programmable weights or synapses. Our brain essentially operates as in-memory computing paradigm which enables it to be highly parallel circumventing the von Neumann bottleneck inherent in conventional digital computers. Building large scale neuromorphic systems using traditional CMOS is inefficient since it takes roughly ten transistors for a synapse and even more for building a neuron with rudimentary functionality. Given our brain has ~$10^{11}$ neurons and ~$10^{14-15}$ synapses, building a brain-like consuming machine with CMOS transistors will be equivalent to making millions of state-of-the-art chips for emulating a single brain. Hence, scientists have been exploring emerging devices with tiny form factors and energy consumption that has intrinsic properties suitable for emulating synaptic and neuronal functionality for building large scale brain inspired machined on chip. Comprehensive reviews on emerging devices for such applications can be found elsewere \cite{zhu2020comprehensive, islam2019device, ielmini2019emerging, wang2020resistive}. Here, as an illustrative example, we discuss memristors, which have garnered a lot of attention in this domain during last 15 years since the first experimental demonstration in 2008 \cite{strukov2008missing}.

\begin{figure}
    \centering
    \includegraphics[width=0.9\textwidth]{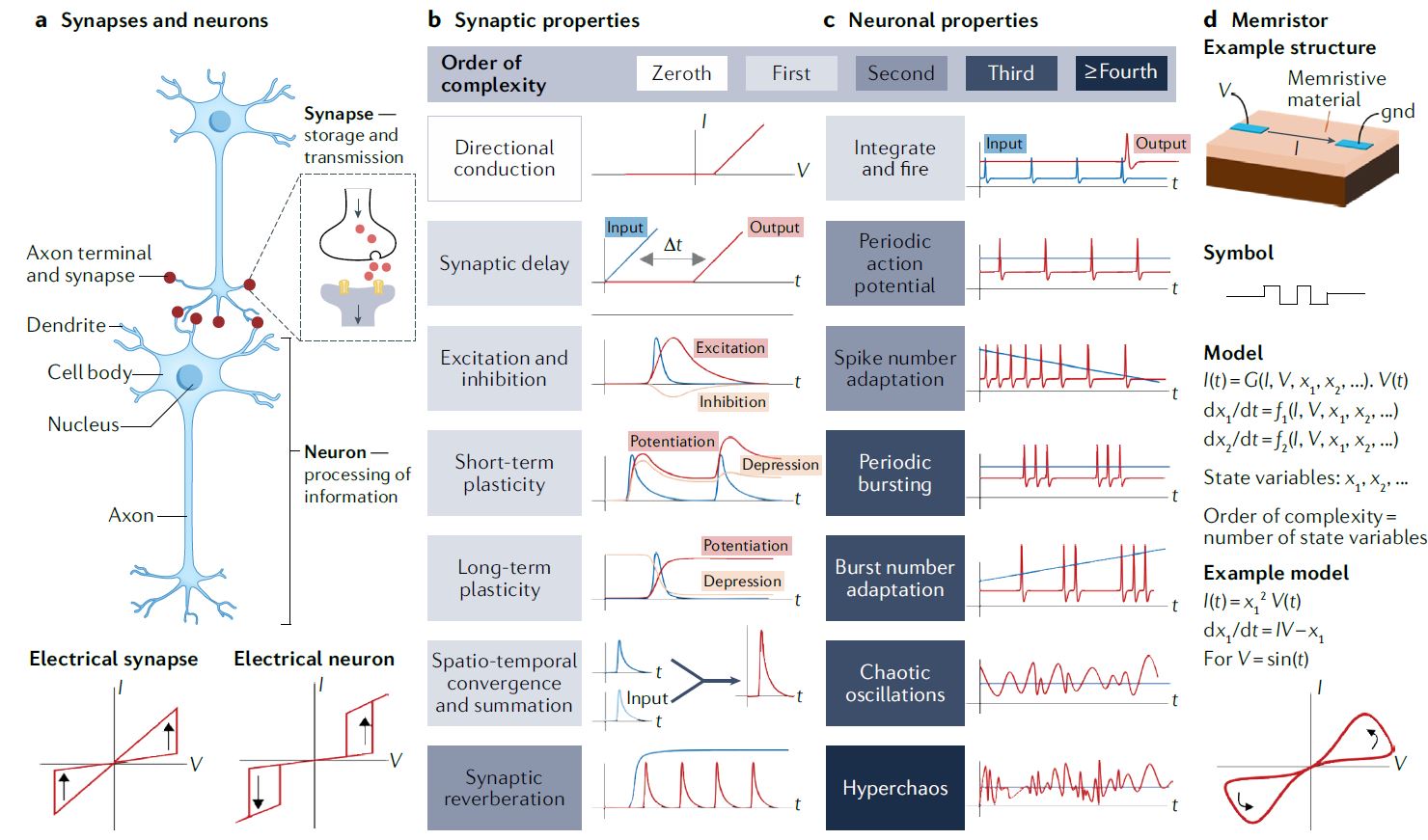}
    \caption{Concept of memristor and memristive behaviours of various complexity. (a) Illustrations of a biological neuron and synapses, along with the current–voltage characteristics of synaptic and neuronal electrical devices. (b,c) Examples of synaptic (panel b) and neuronal (panel c) behaviours that require different orders of complexity. (d) | Illustration of a memristor and its basic model, depicting with an example how state variables connect currents and voltages with a temporal history dependence. $\Delta t$, time delay; G, conductance; gnd, electrical ground; I, current; t, time; V, voltage\cite{kumar2022dynamical}. Reproduced with permission from Kumar et al., Nature Reviews Materials 7.7 (2022). Copyright  2022 Springer Nature.}
    \label{fig:complexity}
\end{figure}

Most efforts in bio-inspired computing thus far have focused on mimicking primitive lower-order biological complexities. Historically, such complexities are emulated using transistor-based circuits (such as central and graphics processing units, CPUs and GPUs) to simulate multiple dynamical equations \cite{nawrocki2016mini, he2021recent} but recent advances in memristors have made this approach easier. Memristors, predicted in 1971 \cite{chua1971memristor} and connected to physical devices in 2008 \cite{strukov2008missing},  are electrical circuit elements that embody at least one state equation (differential equation of the state variable with respect to time) and, thus, at least first-order complexity (Figure \ref{fig:complexity}(d)).
The incorporation of state equations necessarily leads to history-dependent behaviours in the current–voltage plot, in either volatile (memory disappears at zero bias) or non-volatile (memory is retained at zero bias) form. Processes such as temperature-driven Mott transitions \cite{pickett2013scalable}
and field-driven defect generation and recombination \cite{zhu2019ovonic} lead to volatile or sometimes partially volatile memory effects \cite{wang2017memristors}. By contrast, processes such as electrochemical defect migration \cite{kumar2016direct}, spin injection \cite{pershin2008spin}, \cite{manipatruni2018beyond}, ferroelectric or ferromagnetic switching \cite{endoh2016overview}, and crystalline–amorphous phase transitions \cite{raoux2014phase} lead to non-volatile memory effects and are all captured within the memristor framework.

Here, we provide several examples of synaptic and neuronal memristors of different orders of complexity along with their working mechanisms. Several recent reviews have comprehensively covered memristive switching materials, mechanisms, and device-level performance \cite{wang2019overview,wang2020resistive, hong2018oxide},  including special material classes, such as 2D materials \cite{rehman2020decade,feng20202d}, Mott insulators \cite{wang2019mott}, organic materials \cite{goswami2020organic}, and carbon nanomaterials \cite{ahn2018carbon}. Here, instead, we focus on higher-complexity memristive materials and devices and discuss how complex computing can be achieved by taking advantage of the intrinsic device dynamics. A few representative examples that illustrate complexity beyond the simple (and often static) functions that can be engineered in memristive electronic devices are summarized in Figure \ref{fig:complexity}.


\subsubsection{Synapse}

The non-volatile resistance switching in the first experimentally demonstrated memristor \cite{strukov2008missing} was caused by the movement of oxygen vacancies in TiO$_2$ under the influence of an electric field. It was a first-order memristor  in the sense that the dynamics of the underlying state variable (width of the TiO$_2$ region occupied by the oxygen defects)  can be modeled using a first-order differential equation. Most memristors built so far are first-order, that is, one dominant dynamical process that enables memory. Later, memristors based on TaO$_x$(\cite{kim2015experimental}) were designed to be sensitive to thermal effects. These are second-order devices with two state variables, namely, the radius of the filament of oxygen vacancies and temperature. The device exhibited dynamic volatile memory along with the static non-volatile resistance switching property. Second-order effects were also seen in HfO$_2$ memristors  \cite{rodriguez2018characterization}. Another second-order memristor dynamics was found in ferroelectric memristors \cite{mikheev2019ferroelectric} with built-in electric fields and the dynamical response of the interfacial defects being the two state variables. These devices also exhibit temporal memory along with the resistance switching behaviour. In addition to solid state devices, biomolecular volatile memristors with SRDP (spike rate-dependent plasticity) have also been reported \cite{najem2018memristive}. These devices are even more biologically faithful in their composition, ion-channel based conduction, and native time scale of operation and have been used for solving benchmark problems with promising accuracy \cite{hasan2018biomimetic}. The functionality is governed by two independent first-order dynamics, namely, pore generation, and electrowetting. In addition to emulating chemical synapse, biomolecular memristors have also been used to build artificial electrical synapse  that exhibits voltage-dependent,
dynamic changes in its conductance \cite{koner2019memristive}. 

\subsubsection{Neuron}
First-order neuronal memristors essentially exhibit volatile switching in the current–voltage plane and some simple dynamics (such as a characteristic response time). Volatile switching can be caused by Mott transition, thermal runaway, tunnelling, etc. For example, modelling of threshold switching in TiO$_2$ as a first-order process with internal temperature as the state variable \cite{alexandrov2011current} showed that, as temperature increases due to Joule heating, the superlinear temperature dependence of the conductance makes the conductance increase, which, above a threshold, is a runaway process, leading to volatile (reversible) switching.  Volatile memristors placed in a relaxation oscillator circuit can exhibit self-sustained oscillations. For example, volatile memristors (exhibiting current-controlled negative differential resistance) with a parallel capacitor can exhibit oscillations via two alternating dynamical processes: charging–discharging of the capacitor and volatile switching of the memristor, thus exhibiting second-order complexity. The electrode structure of a NbO$_2$ volatile-switching memristor was shown to form a built-in capacitor, which was sufficient to create oscillations without the need for any external capacitor \cite{pickett2011coexistence}. The device was modelled with a Mott-transition-driven volatile filament (conduction channel) formation process, although later models were based on more realistic and general thermal runaway processes \cite{kumar2020third}.

The only reported third-order memristor \cite{kumar2020third} was constructed using NbO$_2$ and modelled with three state variables: temperature (representing internal thermal dynamics), charge on the built-in capacitor (representing charge dynamics), and the speed of formation of a metallic region (a volatile filament resulting from the Mott transition dynamics). The devices were carefully designed in structure and material stoichiometry to enable all the above dynamics. When powered by a tunable static voltage input, a single device could produce 15 different neuronal dynamics (including spiking, bursting and chaos). Although third- order complexity can produce many key neuronal behaviours, their usefulness in designing computational system is still a open question that needs rigorous examination. In addition to single neuron, dynamic volatile memristors, and memcapacitors with higher-order behavior have recently found application as a reservoir in the reservoir computing (RC) system, replacing a complex recurrent neural network (RNN)  for solving several benchmark temporal, dynamic, and chaotic problems with low cost energy efficient analog hardware implementation \cite{du2017reservoir, moon2019temporal, zhong2022memristor, hossain2023energy}.

\section{State-of-the-art Neuromorphic Processors}
\label{sec_proc}

In this section, we give a brief summary of state-of-the-art neuromorphic processors. Several research groups from academia and industry have reported very promising implementations of neuromorphic processors. For example, a mixed-signal, multi-core neuroprocessor called dynamic neuromorphic
asynchronous processor (DYNAP) was reported, which combines the efficiency of analog computational circuits with the robustness of asynchronous digital logic for communications \cite{moradi2017scalable} and was implemented in 180 nm CMOS process. Thakur et al. introduced an improved version called DYNAP with
scalable and learning devices (Dynap-SEL) containing additional features implemented in a 28 nm FDSOI (Fully Depleted Silicon-On-Insulator) process \cite{thakur2018large}. There are four cores with each containing 16$\times$16 analog neurons where each neuron has 64 programmable (4-bit) synapses. There is an additional fifth core containing 1 × 64 analog neurons, 64$\times$128 plastic synapses (on-line learning capability), and 64$\times$64 programmable
synapses. 

Two other famous neuroprocessors named ``SpiNNaker'' \cite{furber2014spinnaker} and ``BrainScaleS''\cite{schemmel2010wafer} came out of the Human Brain Project (HBP) in Europe \cite{amunts2016human}. The SpiNNaker, developed by researchers at the University of Manchester, contains more than one million parallel ARM processors, which are used to model one billion spiking neurons with biologically realistic synaptic connections in real time \cite{furber2014spinnaker}. On the other hand, BrainScaleS, a mixed-signal neuromorphic system at wafer-scale with upwards of 40 million synapses
and 180 thousand neurons, has been developed from a research collaboration between the University of Heidelberg and the Technische Universität Dresden \cite{schemmel2010wafer}.

TrueNorth, a famous neuroprocessor from IBM \cite{merolla2014million}, consists of 4096 neurosynaptic cores with 1 million digital neurons and 256 million synapses tightly interconnected by an event-driven routing infrastructure consuming 65 mW power. They also introduced a novel hybrid asynchronous–synchronous model along with new CAD tools for the design and verification. Another prominent neuroprocessor is Loihi from Intel \cite{davies2018loihi}, which contains 128 neuromorphic cores, three ×86 processor cores, and four communication interfaces that extend the mesh in four directions to other chips. Each neuromorphic core has 1024 primitive spiking neural units grouped into sets of neuronal trees. The mesh
protocol can support up to 16,384 chips and 4096 on-chip cores using hierarchical addressing. Loihi introduced several novel features, such as hierarchical connectivity, dendritic compartments, synaptic delays, and programmable synaptic learning rules.

A family of dynamically adaptive neural processors has been developed by the TENNLab neuromorphic research group at the University of Tennessee. The first one is called DANNA (dynamic adaptive neural network array) \cite{dean2014dynamic}, which was initially designed for FPGA and later adapted for 130 nm CMOS ASIC (application specific integrated circuit) implementation. An improved version called DANNA2 was introduced in 2018 with improved network density, achievable clock speeds, and training convergence rate \cite{mitchell2018danna}. In parallel, a mixed-signal extension known as memristive dynamic adaptive neural network array (mrDANNA) was later developed that utilized memristor devices in the synapses to improve the efficiency of the neuromorphic system \cite{chakma2017memristive} with an online learning methodology called synchronous digital long-term plasticity (DLTP). Currently, TENNLab is working on developing a convergent and flexible architecture as part of a reconfigurable and very efficient neuromorphic system or RAVENS \cite{foshie2021multi}. 

The description above is not exhaustive by any means and meant as in introduction to the exciting world of hardware implementation of a new computing paradigm. Table \ref{SOTA} lists several neuromorphic processor implementations from the literature along with key aspects and references for further enquiry. 

\begin{table}[H]
  \centering
  \caption{State-of-the-art Neuromorphic Processors}
\scalebox{1.2}{
  \renewcommand{\arraystretch}{1.3}
  
\begin{tabular}{|c|c|c|c|c|}
\hline 
Name & Operation & power/energy & timescale & on-chip learning \tabularnewline
\hline 

Dynap-SEL \cite{thakur2018large} & Mixed & $260$ pJ/spike & ns & STDP \tabularnewline
\hline 

FPAA \cite{hasler2019large} & Analog & $< 1 \mu W$ & ms to s & STDP \tabularnewline
\hline 

BrainScaleS\cite{schemmel2010wafer} & Digital & $10$ pJ/transmit & ns & Configurable plasticity \tabularnewline
\hline 

TrueNorth\cite{merolla2014million} & Digital & $60$ mW & ns & none \tabularnewline
\hline 

SpiNNaker\cite{furber2014spinnaker} & Digital & $100$ nJ/neuron + $43$ nJ/synapse & ns & Configurable \tabularnewline
\hline 

mrDANNA\cite{chakma2017memristive} & Mixed & $22.31$ pJ/neuron/spike + $0.48$ pJ/synapse/spike & ns to $\mu s$ & DLTP \tabularnewline
\hline 

Loihi\cite{davies2018loihi} & Digital & $81$ pJ/neuron + $120$ pJ/synapse & ns & Configurable STDP \tabularnewline
\hline

\end{tabular}
}
\label{SOTA}
\end{table}

\section {Algorithm and Architecture} 
\label{sec_alg}
In this chapter, we focus on neuromorphic systems that implement SNNs. A key ongoing challenge in the field of SNNs is how to do training or learning effectively.  Here, we overview some of the common approaches used in training SNNs for vision applications. It is worth noting, many of these algorithms have been evaluated on simple image classification tasks such as MNIST and CIFAR-10, with a few extending on to ImageNet-style classifications.

\subsection{Mapping Traditional ANNs to SNNs}

\begin{figure}
    \centering
    \includegraphics[width=0.73\textwidth]{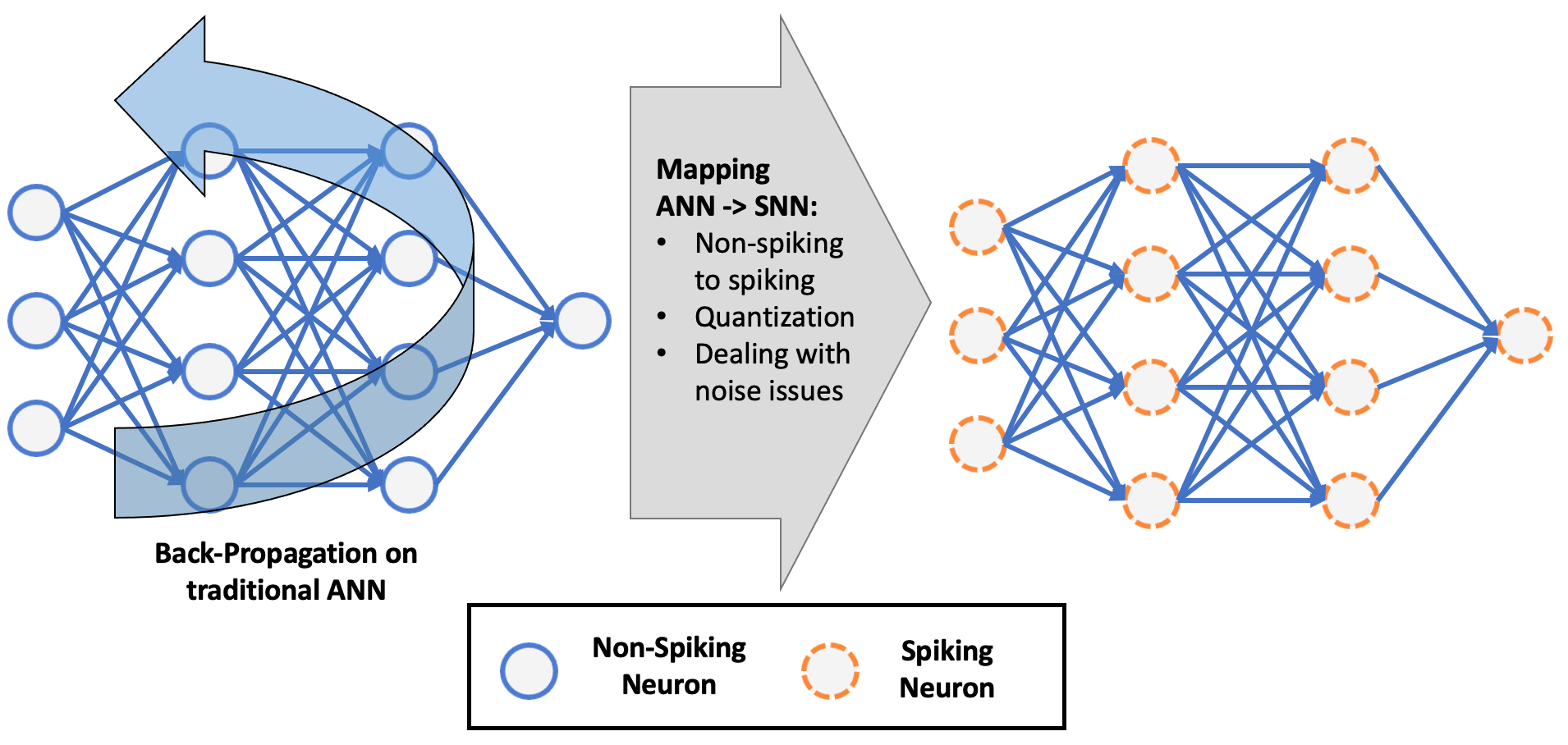}
    \caption{Mapping procedures take a pre-trained ANN and map it into an SNN suitable for neuromorphic hardware.}
    \label{fig:mapping}
\end{figure}

One of the common approaches used in the field is to train a traditional ANN and then create a mapping from that ANN to an SNN for neuromorphic hardware deployment~\cite{diehl2015fast,diehl2016conversion,hunsberger2016training,sengupta2019going,rueckauer2017conversion,neil2016learning,esser2015backpropagation,cao2015spiking} (Figure \ref{fig:mapping}). In creating a mapping from a traditional ANN to an SNN on hardware, several issues are encountered.  First, some accommodation has to be made to move from a nonlinear activation function like a rectified linear unit (ReLU) or sigmoid to a spiking neuron model. Second, due to hardware limitations, there may be restricted precision on the synaptic weight values, requiring quantization of the weight values.  Many deep learning software packages now allow for quantization as part of the training process, but the quantization level required for the hardware must be known a priori to leverage this (i.e., it must be known which hardware system will be targeted). Finally, mapping onto neuromorphic systems with emerging architectures may also require dealing with cycle-to-cycle and device-to-device variation, which can add noise to the operation of the network. Any of these issues may lead to a performance degradation from the original ANN to the SNN on neuromorphic hardware.

\subsection{Spike-Based Quasi-Backpropagation}

\begin{figure}
    \centering
    \includegraphics[width=0.32\textwidth]{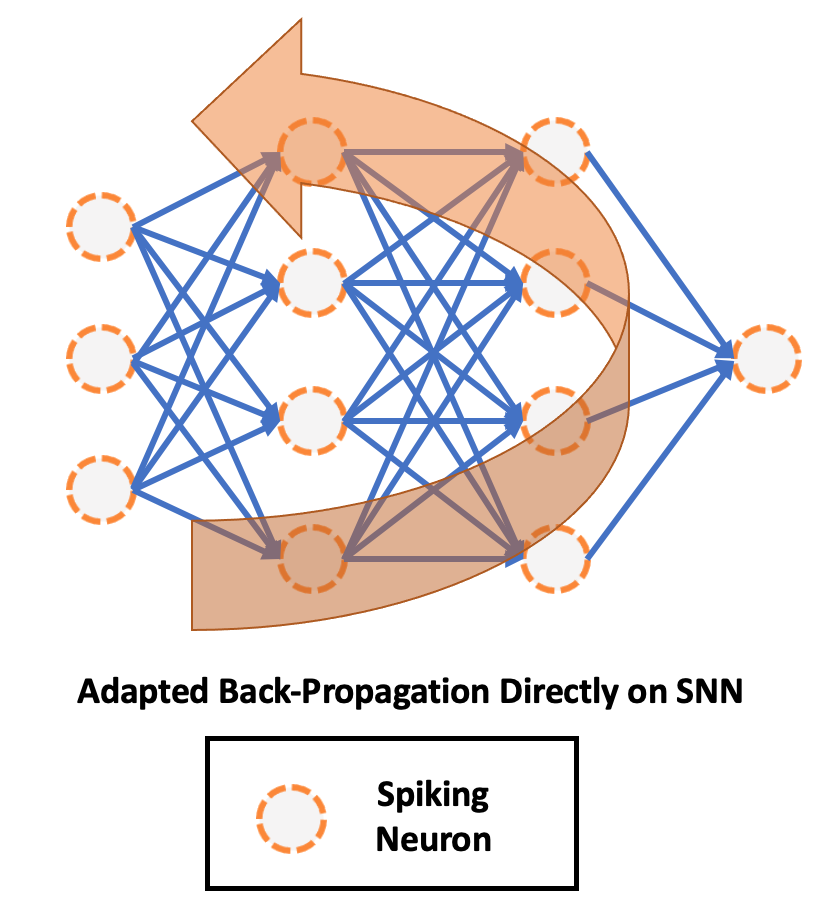}
    \caption{Spike-based quasi-backpropagation approaches adapt backpropagation to work directly on SNNs.}
    \label{fig:quasibp}
\end{figure}

Another approach that is taken in training SNNs for neuromorphic hardware is to directly adapt the training procedure to produce an SNN rather than an ANN.  In this case, the typical training procedures of backpropagation or stochastic gradient descent are modified so that they will produce an SNN suitable for hardware deployment (Figure \ref{fig:quasibp}).  SpikeProp pioneered a bespoke gradient descent approach for spiking neural networks, specifically focused on first-spike times as a proxy for the neuron's output value and applying a process similar to traditional gradient descent~\cite{bohte2000spikeprop}. 

In~\cite{severa2019training}, an approach for training binary-activated networks suitable for hardware deployment is presented.  This approach, called Whetstone, begins with differentiable activation functions such as sigmoid or ReLU and gradually ``sharpens" those functions over the course of training to behave like binary activation functions that are suitable for neuromorphic deployment.  

Several approaches have adapted back-propagation by using a surrogate gradient. Wu et al. introduce an approach for training both the spatial and temporal aspects of SNNs~\cite{wu2018spatio}. They adapt for traditional back-propagation by maintaining an approximated derivative of spike activity throughout. Neftci et al. provide an overview of different approaches for using surrogate gradients to perform backpropagation-based training in recurrent SNNs~\cite{neftci2019surrogate}. Lee et al. treat the membrane potential of the spiking neurons as differentiable signals, which allows the application of backpropagation~\cite{lee2016training}. Others have adapted backpropagation through time (BPTT) to be amenable to recurrent SNN architectures~\cite{bellec2018long}. Still yet, others have used differentiation on the spike representation to overcome the non-differentiability issue~\cite{meng2022training}.  An overview of other deep learning-like training approaches for SNNs is provided in~\cite{tavanaei2019deep}. 

\subsection{Plasticity-Based Training}

\begin{figure}
    \centering
    \includegraphics[width=0.55\textwidth]{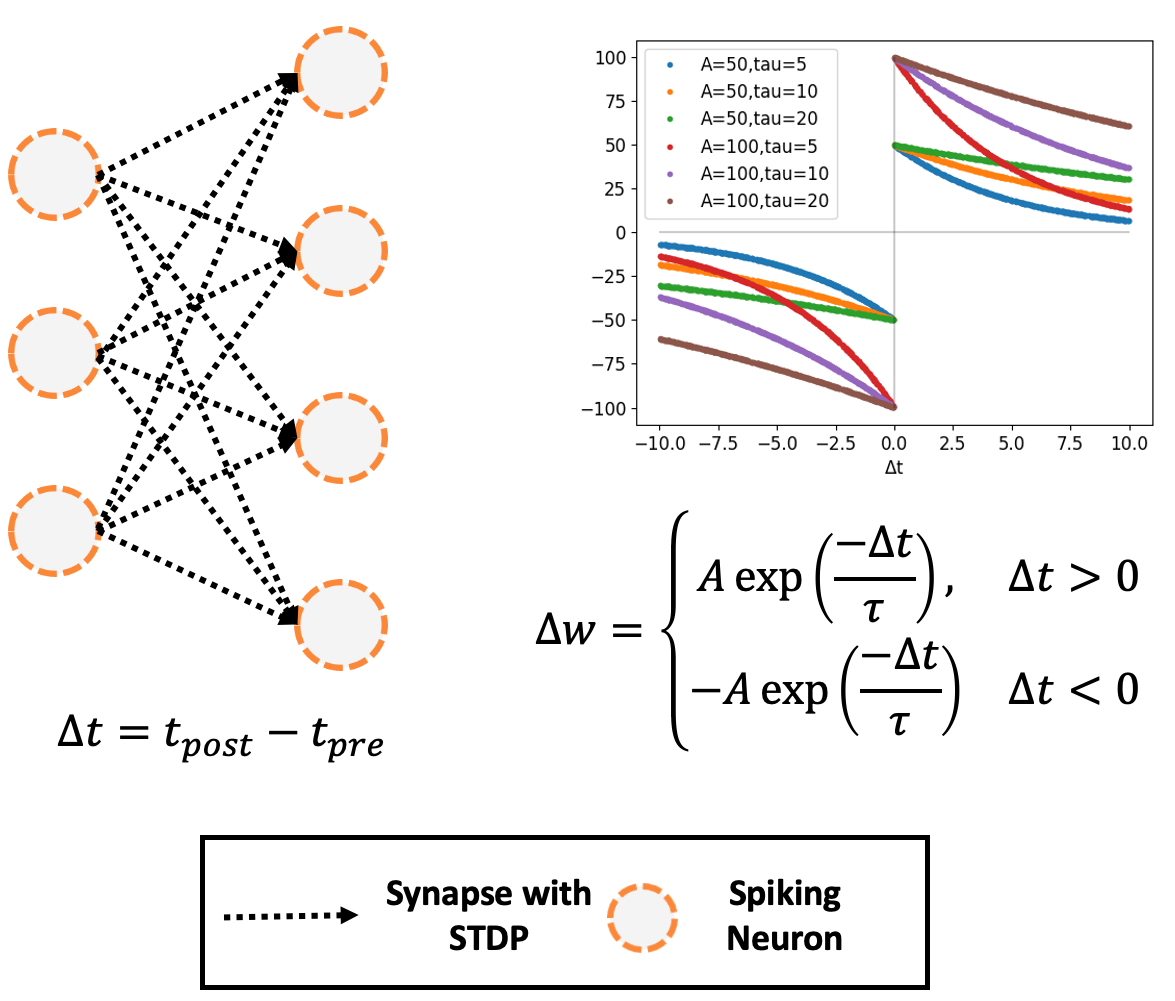}
    \caption{One common example of synaptic plasticity mechanisms is STDP.  In this figure, a simple STDP equation is given, along with plots showing how the weight changes are affected by the parameter values.}
    \label{fig:stdp}
\end{figure}

Spike-timing-dependent plasticity (STDP) is a synaptic-weight plasticity mechanism in which the weight of the synapse is adjusted based on the relative spike times of the pre- and post-synaptic neuron (Figure \ref{fig:stdp}). If the post-neuron spikes after the pre-neuron, it leads to potentiation or, increase in the synaptic weight. Conversely, if the post-neruron fires before the pre-neuron, it causes depression or, decrease in the synaptic weight. The weight change usually decays exponentially with respect to time interval between two spikes as shown in  Figure \ref{fig:stdp} and A and $\tau$ are learning rates and time constants governing the weight change, $\Delta w$. Here, we show symmetric behavior for simplicity but in general, the parameters can be different for potentiation and depression.  It has been observed in biological neural systems~\cite{caporale2008spike}, and it has been implemented in a vast array of neuromorphic hardware implementations~\cite{schuman2017survey}. STDP has been demonstrated as a standalone unsupervised learning rule for SNNs on simple vision tasks, such as MNIST classification~\cite{diehl2015unsupervised}, but STDP alone has been shown to have difficulties scaling to deeper architectures and more complicated tasks. STDP has also been used in combination with traditional learning methods, for example, using STDP for training convolutional filters and developing a supervised STDP-based learning rule that is meant to approximate gradient descent~\cite{tavanaei2018training}. Others have used a layer-by-layer STDP training approach to leverage STDP in deeper networks for more complicated tasks~\cite{kheradpisheh2018stdp}. 

\subsection{Reservoir Computing (RC)}

\begin{figure}
    \centering
    \includegraphics[width=0.57\textwidth]{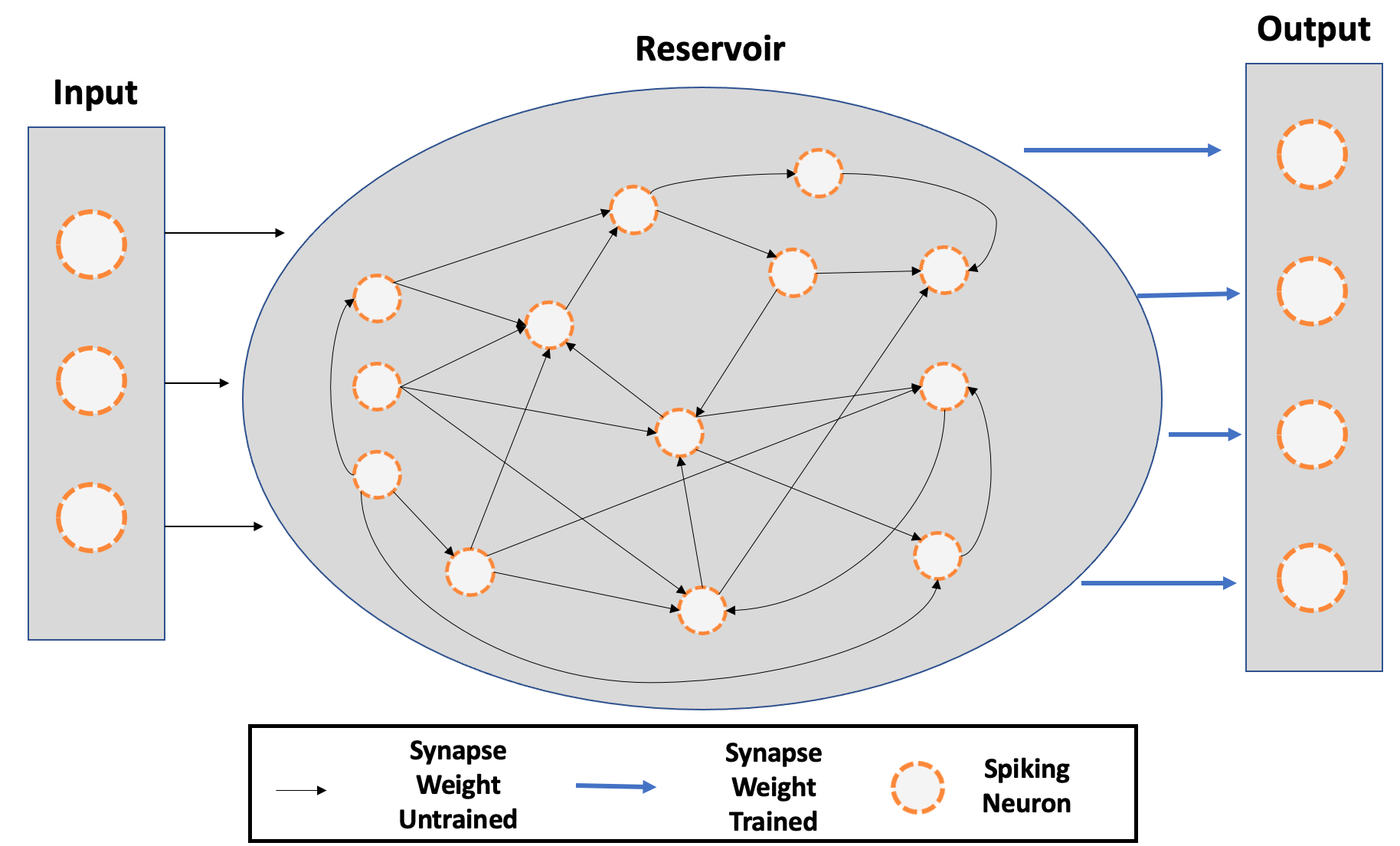}
    \caption{Reservoir computing or liquid state machines is commonly used as an algorithm for training SNNs. In this case, only the weights between the reservoir (or liquid) and the outputs are trained.}
    \label{fig:reservoir}
\end{figure}

In RC, a sparse, highly recurrent neural network is defined as the ``reservoir", and data is passed through the reservoir and then processed by a trained, typically linear ``readout" layer that interprets the output of the reservoir (Figure \ref{fig:reservoir}). Reservoirs are typically well-suited to temporal data tasks.  In the context of neuromorphic computing, reservoirs were typically implemented using a recurrent spiking neural network; in this case, the RC approach is referred to as liquid state machines. RC approaches have also been applied to static computer vision tasks as well as event-based camera data and videos. A key issue with RC is defining the appropriate structure of the reservoir. Iranmehr, et al., have demonstrated a spiking reservoir approach on N-MNIST, an event-based version of the MNIST dataset, where they define the reservoir based on ionic density in an ionic environment~\cite{iranmehr2020ils}. Others have used evolutionary approaches to define the reservoir structure~\cite{reynolds2019intelligent, tang2022evolutionary}. Others have proposed leveraging ensembles of reservoirs to achieve higher accuracy on image classification tasks~\cite{wijesinghe2019analysis}. 

The role of the reservoir in RC is to nonlinearly transform sequential inputs into a high-dimensional space such that the features of the inputs can be efficiently read out by a simple learning algorithm. Therefore, instead of RNNs, other nonlinear dynamical systems can also be used as reservoirs. In particular, physical RC using reservoirs based on physical phenomena has recently attracted increasing interest in many research areas. Various physical systems, substrates, and devices have been proposed including electronic \cite{soriano2014delay, schrauwen2008compact, zhang2015digital, soures2017robustness}, photonic \cite{van2017advances, brunner2018tutorial}, mechanical \cite{hauser2011towards} \cite{caluwaerts2014design}, biological \cite{dranias2013short, sussillo2009generating, jones2007there}, quantum \cite{fujii2017harnessing} and spintronic \cite{torrejon2017neuromorphic, bourianoff2018potential} RC. A motivation for physical implementation of reservoirs is to realize compact and fast information processing devices with low learning cost. 

There are several requirements for a physical reservoir to efficiently solve computational tasks \cite{tanaka2019recent}. (1) High dimensionality is necessary to map inputs into a high-dimensional space. This property facilitates the separation of originally inseparable inputs in classification tasks and allows reading out spatiotemporal dependencies of inputs in prediction tasks. The dimensionality is related to the number of independent signals obtained from the reservoir. (2) Nonlinearity is necessary for a reservoir to operate as a nonlinear mapping. This property allows inputs that are not linearly separable to be transformed into those that are linearly separable in classification tasks. It is also useful for effectively extracting nonlinear dependencies of inputs in prediction tasks. (3) Fading memory (or short-term memory) \cite{maass2004fading} is necessary to ensure that the reservoir state is dependent on recent-past inputs but independent of distant-past inputs. It is also referred to as the echo state property, indicating that the influence of past inputs on the current reservoir states and outputs asymptotically fades out \cite{yildiz2012re}. Such a property is particularly important for representing sequential data with short-term dependencies. (4) Separation property is required to separate the responses of a reservoir to distinct signals into different classes. On the other hand, a reservoir should be insensitive to unessential small fluctuations, such as noise, so that similar inputs are classified into the same class.

\subsection{Evolutionary Algorithms}

To deal with the highly recurrent and spiking nature of recurrent SNNs, others have turned to evolutionary algorithms to evolve the structure and/or parameters of the networks~\cite{schuman2020evolutionary, schuman2016evolutionary}. Though evolutionary approaches have shown to perform well on a variety of tasks and to outperform deep learning-style SNN training on imitation learning tasks for control~\cite{schuman2022evolutionary}, there have been limited results on computer vision tasks.  We expect that the use of evolutionary algorithms for computer vision tasks for SNNs would be more suited towards neural architecture search~\cite{liu2021survey} and used alongside one of the other training approaches described above. 


\section{Application} 
\label{sec_app}
SNNs are promising for analyzing both static and spatio-temporally varying sequential data. In this section, we discuss three promising applications of SNNs, namely static image classification, neuromorphic dataset classification, and dancing pose estimation.

\subsection{Static Image Classification}

\begin{table}
  \centering
  \caption{Static Image Classification with SNN}
  \scalebox{1}{
  \renewcommand{\arraystretch}{1.3}
  
\begin{tabular}{|c|c|c|c|c|c|c|c|}
\hline 
Problem &\textbf{Paper} & Year & Neuron & Input Coding & Algorithm & Architecture & Accuracy (\%) \tabularnewline
\cline{1-8}
\cline{1-8}
\multirow{5}{*}{MNIST}&\textbf{\cite{diehl2015unsupervised}} &2015 & LIF & Rate & STDP & 2FC & 95  \tabularnewline
\cline{2-8}
&\textbf{\cite{mostafa2017fast}} &2017 & IF & Temporal & Backprop & 784FC-600FC-10FC & 96.98  \tabularnewline
\cline{2-8}
&\textbf{\cite{srinivasan2019restocnet}} &2019 & LIF & Rate & Stochastic STDP & 36C3-2P-128FC-10FC & 98.54  \tabularnewline
\cline{2-8}
&\textbf{\cite{zhang2020efficient}} &2020 & IF & Temporal & ANN-SNN & 32C5-P2-64C5-P2-1024FC-10FC & 99.41  \tabularnewline
\cline{2-8}
&\textbf{\cite{zhang2020temporal}} &2020 & LIF & Encoding Layer & Backprop & 15C5-P2-40C5-P2-300FC & 99.53  \tabularnewline
\cline{1-8}
\cline{1-8}
\multirow{6}{*}{CIFAR-10}&\textbf{\cite{rueckauer2017conversion}} &2017 & IF & Rate & ANN-SNN & 4 Conv, 2 FC &90.85 \tabularnewline
\cline{2-8}
&\textbf{\cite{sengupta2019going}} &2019 &IF &Rate &ANN-SNN & VGG16 &91.55  \tabularnewline
\cline{2-8}
&\textbf{\cite{rathi2020enabling}} &2020 & LIF & Rate & Hybrid & VGG16 &92.02  \tabularnewline
\cline{2-8}
&\textbf{\cite{han2020deep}} &2020 & IF &Temporal & ANN-SNN & VGG16 & 93.63  \tabularnewline
\cline{2-8}
&\textbf{\cite{deng2022temporal}} &2022 & LIF & Rate & TET & ResNet-19 & 94.50  \tabularnewline
\cline{2-8}

&\textbf{\cite{duan2022temporal}} &2022 & LIF & Rate & Surrogate Gradient & ResNet-19 & 95.60 \tabularnewline
\cline{1-8}
\cline{1-8}

\multirow{6}{*}{CIFAR-100}&\textbf{\cite{han2020rmp}} &2020 & RMP(soft-reset) & Rate & ANN-SNN & ResNet-20 & 67.82  \tabularnewline
\cline{2-8}
&\textbf{\cite{li2021differentiable}} &2021 &LIF & Rate & Surrogate Gradient & ResNet-18 & 74.24  \tabularnewline
\cline{2-8}

&\textbf{\cite{deng2022temporal}} & 2022 & LIF & Rate & TET & ResNet-19 & 74.72  \tabularnewline
\cline{2-8}

&\textbf{\cite{duan2022temporal}} &2022 & LIF & Rate & Surrogate Gradient & ResNet-19 & 78.76  \tabularnewline
\cline{1-8}
\cline{1-8}

\multirow{6}{*}{ImageNet}&\textbf{\cite{sengupta2019going}} &2019 & IF & Rate & ANN-SNN & VGG16 & 69.96  \tabularnewline
\cline{2-8}
&\textbf{\cite{rathi2020enabling}} &2020 & LIF & Rate & Hybrid & VGG16 & 65.19  \tabularnewline
\cline{2-8}
&\textbf{\cite{han2020deep}} &2020 & IF & Temporal & ANN-SNN & VGG16 & 73.46  \tabularnewline
\cline{2-8}
&\textbf{\cite{li2021free}} &2021 & IF & Rate & ANN-SNN & ResNet-34 & 73.45  \tabularnewline
\cline{2-8}

&\textbf{\cite{deng2022temporal}} & 2022 & LIF & Rate & TET & SEW-ResNet-34 & 68  \tabularnewline
\cline{2-8}

&\textbf{\cite{duan2022temporal}} & 2022 & LIF & Rate & Surrogate Gradient & SEW-ResNet-34 & 68.28  \tabularnewline
\cline{2-8}

&\textbf{\cite{yao2023attention}} & 2023 & LIF & Rate & Direct Training & ResNet-104 & 77.08  \tabularnewline
\cline{1-8}

\end{tabular}}
\label{table_static_data}
\end{table}

Researchers have used SNNs on several benchmark image classification problems as summarized by Rathi et al. \cite{rathi2023exploring}. Table \ref{table_static_data} expands upon the prior work to provide a thorough comparison among the performance of various reported recent SNN models on image classification tasks from frame-based static image datasets such as MNIST \cite{lecun1998gradient}, CIFAR10 \cite{krizhevsky2009learning}, CIFAR100 \cite{krizhevsky2009learning}, and ImageNet \cite{deng2009imagenet}). Of course, conventional deep ANNs have performed very well in these problems, but as can be seen from this table, innovative SNN implementation has become quite competitive in this domain over the years.

\subsection{Neuromorphic Dataset (from event-based camera) Classification}

Neuromorphic cameras \cite{lichtsteiner2008128}, also known as Dynamic Vision Sensors (DVS) or event cameras, have a silicon retina design based on mammalian vision, making them sensitive to moving targets and fluctuating lighting conditions. Each pixel operates asynchronously, independently monitoring logarithmic minute brightness changes, ensuring sensitivity to motion in various lighting conditions. This mechanism inherently filters out static backgrounds, transmitting detailed contents only when an event occurs. Neuromorphic cameras are resilient in situations ranging from nighttime to glaring noon and exhibit reduced sensitivity to skin color and brightness changes \cite{4541871,gallego2020event}. The can offer very high temporal resolution and low latency (order of microseconds), high dynamic range (140 dB versus 60 dB of standard camera), and low power consumption. 

Since the first commercial event camera of 2008 \cite{lichtsteiner2008128}, there has been a lot of interest in recent years on developing new sensors \cite{posch2010qvga, brandli2014240, finateu20201280x720,son20174,suh20201280, chen2019live, insight} for diverse  applications and algorithms for efficient processing the data from these cameras. According to a recent survey \cite{gallego2020event}, applications of event-based cameras may include real-time interaction systems such as robotics and wearable electronics \cite{delbruck2016neuromorophic}, systems requiring low latency and power in uncertain lighting condition \cite{liu2019event}, object tracking \cite{delbruck2013robotic,glover2016event}, surveillance and monitoring \cite{litzenberger2006estimation}, object/gesture recognition \cite{ lee2014real,orchard2015hfirst}, depth esitmation \cite{rogister2011asynchronous, rebecq2018emvs}, structured light 3D scanning \cite{matsuda2015mc3d}, optical flow estimation \cite{benosman2013event}, HDR (high dynamic range) image reconstruction \cite{Kim2014SimultaneousMA, rebecq2019high}, simultaneous Localization and
Mapping (SLAM) \cite{kim2016real, vidal2018ultimate}, image deblurring \cite{pan2019bringing}, or star tracking \cite{chin2019star}. Since, they asynchronously measure per-pixel brightness changes ('events') in contrast to standard cameras measuring absolute brightness at a constant rate, novel methods are required to process their output. Due to its similarity to biological vision and spiking output, we envision that diverse applications related to event-camera would benefit greatly from advances in neuromorphic computing \cite{roy2019towards, gallego2020event}. 

In Table \ref{table_neuro_data}, we are showing SNN results from recent works for neuromorphic datasets such as N-MNIST \cite{orchard2015converting}, CIFAR10-DVS \cite{li2017cifar10}, DVS128 Gesture \cite{amir2017low}, and DVS128 Gait \cite{wang2019ev} dataset. N-MNIST is a basic neuromorphic dataset converted from MNIST by a DVS (Dynamic Vision Sensor) camera. Like MNIST, there are 60,000 training samples and 10,000 testing samples in N-MNIST. Each sample in N-MNIST has a pixel size of 34$\times$34, a channel size of two, and a time length of 300 ms. DVS-CIFAR10 is converted from CIFAR10 by a DVS camera, but is more challenging than CIFAR10 due to its larger environmental noise and intra-class variance. DVS-CIFAR10 contains a total of 10,000 samples with 10 labels: ``airplane", ``automobile", ``bird", ``ca" ``deer", ``do" ``frog", ``horse", ``ship", and ``truck". DvsGesture contains 11 gestures: ``hand-clapping", ``right-hand-wave", ``left-hand-wave", ``right-arm-clockwise", ``right-arm-counter-clockwise", ``left-arm-clockwise", ``leftarm-counter-clockwise", ``arm-roll", ``air-drums", ``air-guitar", and ``other gestures". There are 1342 samples from 29 subjects under three types of illumination and the average duration of each gesture is 6 seconds. DVS128 Gesture consists of a series of human hand and arm gestures recorded by a DVS camera. DVS128 Gait dataset has various gaits from 21 volunteers (15 males and 6 females) under two kinds of viewing angles with 4200 recorded samples and the average duration of each gait is 4.4 seconds. With the development of more mature DVSs, researchers are now developing benchmarks for  neuromorphic dataset and evaluation metrics. We believe that the advantage of SNN over ANN will be more prominent for such datasets, which seem natively suited for this computational paradigm in contrast to traditional frame-based static datasets.

\begin{table}
  \centering
  \caption{Neuromorphic Dataset Classification with SNN}
  \scalebox{1}{
  \renewcommand{\arraystretch}{1.2}
  
\begin{tabular}{|c|c|c|c|c|c|c|}
\hline 
Problem &\textbf{Paper} & Year & Neuron & Algorithm & Architecture & Accuracy (\%) \tabularnewline
\cline{1-7}
\cline{1-7}
\multirow{5}{*}{N-MNIST}&\textbf{\cite{jin2018hybrid}} &2018 &SRM &Backprop &2FC  &98.88   \tabularnewline
\cline{2-7}
&\textbf{\cite{shrestha2018slayer}} &2018 & SRM & Backprop & 12C5-2P-64C5-2P-10FC & 99.20  \tabularnewline
\cline{2-7}
&\textbf{\cite{wu2019direct}} &2019 & LIF & Surrogate Gradient & 128C3-128C3-AP2-128C3-256C3-AP2-
1024FC-Voting & 99.53  \tabularnewline
\cline{2-7}
&\textbf{\cite{deng2020rethinking}} &2020 &LIF &STBP & 128C3-128C3-AP2 -128C3-256C3-AP2-1024FC-10 & 99.42  \tabularnewline
\cline{2-7}
&\textbf{\cite{wu2021tandem}} &2021 & IF & Surrogate Gradient & 5 conv, 2 FC & 99.31  \tabularnewline
&\textbf{\cite{wu2021tandem}} &2021 & LIF & Surrogate Gradient & 5 conv, 2 FC & 99.22  \tabularnewline
\cline{2-7}
&\textbf{\cite{qiao2023batch}} &2023 & IF & Surrogate Gradient & 2 conv, 2linear &99.44  \tabularnewline

\cline{1-7}
\cline{1-7}
\multirow{7}{*}{DVS-CIFAR10}&\textbf{\cite{wu2019direct}} &2019 & LIF & Surrogate Gradient & 128C3-128C3-AP2-128C3-256C3-AP2-
1024FC-Voting & 60.50  \tabularnewline
\cline{2-7}
&\textbf{\cite{kugele2020efficient}} &2020 & IF & ANN-SNN &  4 Conv, 2 FC & 65.61  \tabularnewline
\cline{2-7}
&\textbf{\cite{kim2021revisiting}} &2021 & LIF & Surrogate Gradient & 5 Conv, 3 FC & 63.20  \tabularnewline
\cline{2-7}
&\textbf{\cite{wu2021tandem}} &2021 & IF & Surrogate Gradient & 5 Conv, 2 FC & 65.59  \tabularnewline

&\textbf{\cite{wu2021tandem}} &2021 & LIF & Surrogate Gradient & 5 Conv, 2 FC & 63.73  \tabularnewline
\cline{2-7}
&\textbf{\cite{li2021differentiable}} &2021 & \textbf{LIF} & Surrogate Gradient & ResNet-18 & 75.40  \tabularnewline
\cline{2-7}
&\textbf{\cite{deng2022temporal}} &2022 & LIF  & TET & VGGSNN &83.17  \tabularnewline
\cline{2-7}

&\textbf{\cite{duan2022temporal}} &2022  & LIF  & Surrogate Gradient & VGGSNN &84.90 
\tabularnewline

\cline{1-7}
\cline{1-7}

\multirow{5}{*}{DVS128 Gesture}&\textbf{\cite{amir2017low}} &2017 & LIF & ANN-SNN & 16-layer SNN & 94.59  \tabularnewline
\cline{2-7}
&\textbf{\cite{shrestha2018slayer}} &2018 & SRM  & Surrogate Gradient & 8-layer-SNN & 93.64  \tabularnewline
\cline{2-7}
&\textbf{\cite{fang2021deep}} &2021 & PLIF & Surrogate Gradient & 7B-Net & 97.92  \tabularnewline
\cline{2-7}

&\textbf{\cite{wu2021liaf}} &2021 & LIAF & BPTT & Conv-LIAF & 97.56  \tabularnewline
\cline{2-7}

&\textbf{\cite{yao2021temporal}} &2021 & LIF  & STBP and BPTT & Input-MP4-64C3-128C3-
AP2-128C3-AP2-256FC-11 & 98.61  \tabularnewline

\cline{1-7}
\cline{1-7}

\multirow{5}{*}{DVS128 Gait}&\textbf{\cite{wang2019ev}} &2019 & NA  & BP & 6 CNN, 3 FC & 89.9  \tabularnewline
\cline{2-7}
&\textbf{\cite{wang2021event}} &2021 & NA  & BP & 6 CNN, 3 FC & 94.90  \tabularnewline
\cline{2-7}
&\textbf{\cite{yao2021temporal}} &2021 &LIF &STBP and BPTT &Input-MP4-64C3-128C3-
AP2-128C3-AP2-256FC &87.59  \tabularnewline
\cline{2-7}

&\textbf{\cite{fang2021incorporating}} & 2021 & PLIF & Surrogate Gradient & 5 Conv, 3 FC &89.87  \tabularnewline
\cline{2-7}

&\textbf{\cite{yao2023attention}} &2023 & PLIF & Surrogate Gradient (TCSA) & 5 Conv, 3 FC &92.78  \tabularnewline
\cline{1-7}

\end{tabular}}
\label{table_neuro_data}
\end{table}

\subsection{Dancing Pose Estimation}

Technology-mediated dancing (TMD) uses digital systems to enable remote, engaging, and health-promoting dance activities as part of gaming and immersive experiences, increasingly blending digital and physical realities \cite{lopez2013efectividad,cheng2017effects,marquez2017regular}. TMD forms range from gaming console games to Virtual Reality (VR) platforms, integrating with users' living spaces. Human pose estimation (HPE) is critical in TMDs, as it identifies users' unique, complex dance poses for computer interaction. TMD requires high-fidelity HPE that functions reliably in diverse, challenging, and realistic indoor environments, including dynamic lighting and background conditions.

Contemporary HPE systems predominantly rely on depth and RGB cameras \cite{Chen2022AnatomyAware3H,Chen2020MonocularHP,Hassan2019Resolving3H}, which struggle to generate ultra-fast, high-speed pose inferences due to limited frame rates. This limitation is crucial for applications like VR dance games and high-frequency motion characterization for tremor monitoring. Neuromorphic cameras can achieve a bandwidth of over 10 million events per second with low latency, making them suitable for high-frequency inference. RGB-based HPE struggles in low-light conditions and depth cameras have a limited working depth range, while neither camera inherently distinguishes between static and moving objects, leading to a waste of transmission bandwidth. These issues compromise HPE robustness in dynamic settings, and depth cameras also require significant power consumption.

\begin{figure*}
    \centering
    \includegraphics[width=\linewidth]{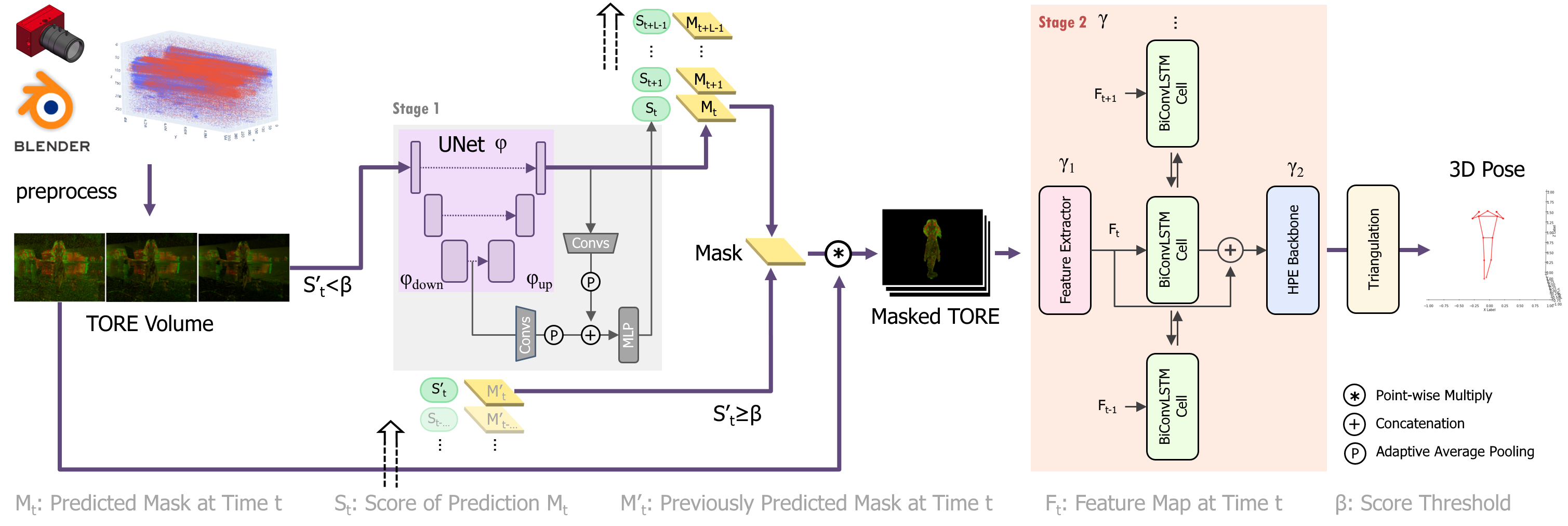}
    \caption{The pipeline of \emph{YeLan}. It initially processes the event stream into TORE (Time-Ordered Recent Event) volumes, which are subsequently sent to the stage one human body mask prediction network. This network predicts a series of masks for the ensuing frames, accompanied by quality-assessment scores to minimize computation costs. The estimated human mask undergoes point-wise multiplication with the original TORE volume before advancing to the next stage. Stage two encompasses the human pose estimation network, where BiConvLSTM and three hourglass-like refinement blocks are employed to estimate the heatmap of joints' projections on three orthogonal planes. The precise 3D coordinates of these joints are determined through a triangulation method based on these heatmaps.}
    \label{fig:pipeline}
\end{figure*}

DVS HPE has gained interest due to its advantages, but current datasets in the field exhibit limitations concerning real-world applicability. They primarily focus on fixed everyday movements and are collected under optimal lighting conditions with static backgrounds, causing models to struggle with intricate movements like those in dance performances. To address this, two novel DVS dancing HPE datasets are introduced: one featuring a real-world dynamic background under various lighting conditions, and another synthetic dataset with variable human models, motion dynamics, clothing styles, and background activities, generated using a comprehensive motion-to-event simulator.

\begin{figure*}
    \centering
    \includegraphics[width=\linewidth]{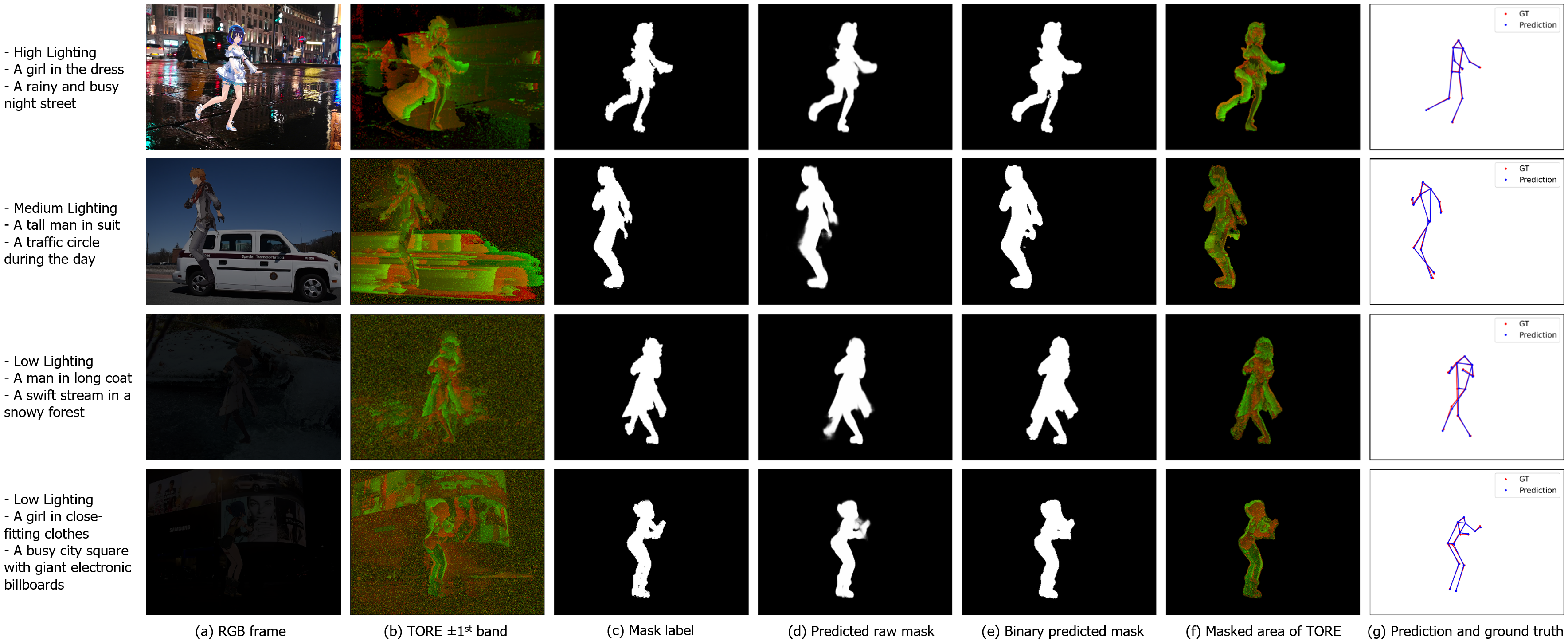}
    \caption{Sample results from \textit{Yelan-Syn-Dataset} in different lighting conditions with dynamic backgrounds. The RGB frames, corresponding event representation TORE, generated masks, ground truth, and predicted 3D human pose are shown.}
    \label{fig:results_with_mask}
\end{figure*}

Previous DVS HPE efforts are also constrained by the ``missing torso" problem, which occurs when neuromorphic cameras only capture moving components and disregard static body parts. To overcome this, a two-stage system called \textit{YeLan} \cite{zhang2023neuromorphic} is proposed (Fig. \ref{fig:pipeline}), which accurately estimates human poses in low-light conditions with noisy backgrounds (Fig. \ref{fig:results_with_mask}). The first stage uses an early-exit-style mask prediction network to eliminate moving background objects, while the second stage employs a BiConvLSTM to facilitate information flow between frames, addressing the missing torso issue. TORE (Time-Ordered Recent Event) volume is also utilized to construct denser input tensors, tackling the low event rate problem in low-light settings. Extensive experiments show that \textit{YeLan} achieves state-of-the-art results on the two proposed new datasets.

\textit{YeLan}, the first neuromorphic camera-based 3D human pose estimation solution designed for dance moves, functions robustly under challenging conditions such as low lighting and occlusion. It effectively overcomes neuromorphic camera limitations while capitalizing on their strengths. An end-to-end simulator was developed, allowing for precise, low-level control over generated events and resulting in the creation of the first and largest neuromorphic camera dataset for dance HPE, called Yelan-Syn-Dataset. This synthetic dataset surpasses existing resources in both quantity and variability. A human subject study was conducted to collect a real-world dance HPE dataset, named Yelan-Real-Dataset, which considers low-light conditions and mobile background content.

\textit{YeLan} is also compared with RGB-based HPE methods, and the results show that although they behave similarly in high-lighting conditions, \textit{YeLan} performs much better in low-lighting conditions. Also, the comparison results show that \textit{YeLan} is less sensitive to dancers' depth, while the performance of RGB-Depth camera-based methods is affected more. Lastly, \textit{YeLan} can be trained on one accumulation frame rate, and operates stably in much higher frame rates, enabling its usage in high-frequency reference. 

To summarize, the work presented discusses existing 3D HPE techniques used in dance games, evaluating their strengths and limitations, and proposes an innovative neuromorphic camera-based approach to address these shortcomings. A real-world dance dataset was collected through human subject studies, and an extensive motion-to-event simulator was constructed to generate a large amount of fully controllable, customizable, and labeled synthetic dance data for pre-training the model. \textit{YeLan} surpasses all baseline models in various challenging scenarios on both datasets. Additionally, in-depth analysis and comparison between different modalities demonstrate \textit{YeLan}'s superiority in many aspects.

\section{Conclusion and Future Directions}
\label{sec_con}
In this chapter, we have provided an overview of the design stack related to spike-based neuromorphic computing for application in computer vision. The confluence of several factors such as slowing down of traditional computing, the exponential increase in time and energy consumption due to growing complexity ANN models in traditional hardware facing progressively larger dataset and the need for ultra-low power computing for millions of smart edge devices calls for an alternative computing paradigm such as neuromorphic computing to address the challenges of today. With the advent of high performance event cameras and neuromorphic imaging, efficient spatio-temporal image and video processing is turning into a multi-disciplinary vibrant field of research spanning from fabricating new materials and devices to developing new learning algorithms informed by neuroscience. 

In recent years, there has been significant progress in SNN algorithms but still scalable supervised and unsupervised learning rules applicable for a wide range of applications remain elusive. Similarly, a host of emerging devices has been reported that exhibit complex dynamic properties to emulate neural activity using intrinsic physical mechanism with tiny form factor and low energy dissipation, yet reliable and robust large scale fabrication with commercially viable yield is still out of reach. More importantly, our knowledge of human perception and cognition along with our mathematical understanding of highly nonlinear large scale complex system such as brain is still at its infancy and as such, nobody is certain which parts of the neural dynamics are beneficial for computation and which parts are historical evolutionary happenstances. Along with computational efficiency, systematic reliability and security framework for ensuring robustness against adversial attack or device variability needs to be developed before deploying these systems for critical applications. In addition, the work in this field so far has focused on neuronal dynamics, although there are dendritic dynamics along with neuroglial cells such as astrocytes, which may play a crucial role in ensuring continual adaptive and flexible lifelong learning and may prove to be indispensable for achieving humanlike artificial general intelligence (AGI), the holy grail of AI. In summary, the high performance and energy efficiency of the human brain at certain perception and cognition tasks waits to replicated in intelligent machines but there are a lot of outstanding challenges and scientific questions that require collaborative research among scientists from multiple disciplines such as computer science, electrical engineering, biology, material science, physics, chemistry, and neuroscience along with significant investment from governments and industries. 

Despite its promise for vision applications, it is important to recognize that several issues remain inadequately addressed. Primarily, a significant proportion of the current literature on DVS are still utilizing accumulated representations, eschewing more efficient and apt SNN or GNN-based methodologies which are potentially more suitable for DVS applications. Additionally, given the relatively nascent status of the DVS-based vision field, both the quantity and quality of available datasets are significantly inferior to those in the conventional frame-based vision domain. Rather than undertaking each data collection process and model training de novo, it would be beneficial to devise superior strategies to adapt these datasets and models effectively to the DVS paradigm. We believe that neuromorphic computing will play a pivotal role in the growth and development of the DVS or event camera for everyday applications. More specifically, spike-based neuromorphic computing and real-time high speed processing of event streams can fundamentally unlock new applications in different domains including scientific computing, climate science, human-computer interaction, and medicine. We anticipate that efficient and robust simulation platform and cross-modality training (e.g., from RGBD to DVS) might play an instrumental role in developing the humanlike AGI for event-based vision.


\IEEEPARstart{}{}

\ifCLASSOPTIONcaptionsoff
  \newpage
  
\fi

\bibliographystyle{IEEEtran}
\bibliography{main.bib}

\begin{thebibliography}{100}
\providecommand{\url}[1]{#1}
\csname url@samestyle\endcsname
\providecommand{\newblock}{\relax}
\providecommand{\bibinfo}[2]{#2}
\providecommand{\BIBentrySTDinterwordspacing}{\spaceskip=0pt\relax}
\providecommand{\BIBentryALTinterwordstretchfactor}{4}
\providecommand{\BIBentryALTinterwordspacing}{\spaceskip=\fontdimen2\font plus
\BIBentryALTinterwordstretchfactor\fontdimen3\font minus \fontdimen4\font\relax}
\providecommand{\BIBforeignlanguage}[2]{{%
\expandafter\ifx\csname l@#1\endcsname\relax
\typeout{** WARNING: IEEEtran.bst: No hyphenation pattern has been}%
\typeout{** loaded for the language `#1'. Using the pattern for}%
\typeout{** the default language instead.}%
\else
\language=\csname l@#1\endcsname
\fi
#2}}
\providecommand{\BIBdecl}{\relax}
\BIBdecl

\bibitem{moore2006cramming}
G.~E. Moore, ``Cramming more components onto integrated circuits, reprinted from electronics, volume 38, number 8, april 19, 1965, pp. 114 ff.'' \emph{IEEE solid-state circuits society newsletter}, vol.~11, no.~3, pp. 33--35, 2006.

\bibitem{dennard1974design}
R.~H. Dennard, F.~H. Gaensslen, H.-N. Yu, V.~L. Rideout, E.~Bassous, and A.~R. LeBlanc, ``Design of ion-implanted mosfet's with very small physical dimensions,'' \emph{IEEE Journal of solid-state circuits}, vol.~9, no.~5, pp. 256--268, 1974.

\bibitem{patterson2016computer}
D.~A. Patterson and J.~L. Hennessy, \emph{Computer organization and design ARM edition: the hardware software interface}.\hskip 1em plus 0.5em minus 0.4em\relax Morgan kaufmann, 2016.

\bibitem{williams2017s}
R.~S. Williams, ``What's next?[the end of moore's law],'' \emph{Computing in Science \& Engineering}, vol.~19, no.~2, pp. 7--13, 2017.

\bibitem{solli2015analog}
D.~R. Solli and B.~Jalali, ``Analog optical computing,'' \emph{Nature Photonics}, vol.~9, no.~11, pp. 704--706, 2015.

\bibitem{preskill2018quantum}
J.~Preskill, ``Quantum computing in the nisq era and beyond,'' \emph{Quantum}, vol.~2, p.~79, 2018.

\bibitem{puaun1998dna}
G.~P{\u{a}}un, G.~Rozenberg, and A.~Salomaa, \emph{DNA computing: new computing paradigms}.\hskip 1em plus 0.5em minus 0.4em\relax Springer, 1998.

\bibitem{mead1990neuromorphic}
C.~Mead, ``Neuromorphic electronic systems,'' \emph{Proceedings of the IEEE}, vol.~78, no.~10, pp. 1629--1636, 1990.

\bibitem{herculano2016human}
S.~Herculano-Houzel, \emph{The human advantage: a new understanding of how our brain became remarkable}.\hskip 1em plus 0.5em minus 0.4em\relax MIT Press, 2016.

\bibitem{mcculloch1943logical}
W.~S. McCulloch and W.~Pitts, ``A logical calculus of the ideas immanent in nervous activity,'' \emph{The bulletin of mathematical biophysics}, vol.~5, pp. 115--133, 1943.

\bibitem{Hebb1949TheTheory}
D.~O. Hebb, ``{The Organization of Behavior; A Neuropsychological Theory},'' \emph{The American Journal of Psychology}, vol.~63, no.~4, p. 633, 1949.

\bibitem{hodgkin1952quantitative}
A.~L. Hodgkin and A.~F. Huxley, ``A quantitative description of membrane current and its application to conduction and excitation in nerve,'' \emph{The Journal of physiology}, vol. 117, no.~4, p. 500, 1952.

\bibitem{fitzhugh1961impulses}
R.~FitzHugh, ``Impulses and physiological states in theoretical models of nerve membrane,'' \emph{Biophysical journal}, vol.~1, no.~6, pp. 445--466, 1961.

\bibitem{nagumo1962active}
J.~Nagumo, S.~Arimoto, and S.~Yoshizawa, ``An active pulse transmission line simulating nerve axon,'' \emph{Proceedings of the IRE}, vol.~50, no.~10, pp. 2061--2070, 1962.

\bibitem{Rosenblatt1958}
F.~Rosenblatt, ``{The Perceptron: A Probabilistic Model for Information Storage and Organization in the Brain},'' \emph{Psychological Review}, vol.~65, no.~6, 1958.

\bibitem{werbos1974beyond}
P.~Werbos, ``Beyond regression: New tools for prediction and analysis in the behavioral sciences,'' \emph{PhD thesis, Committee on Applied Mathematics, Harvard University, Cambridge, MA}, 1974.

\bibitem{rumelhart1986learning}
D.~E. Rumelhart, G.~E. Hinton, and R.~J. Williams, ``Learning representations by back-propagating errors,'' \emph{nature}, vol. 323, no. 6088, pp. 533--536, 1986.

\bibitem{hopfield1982neural}
J.~J. Hopfield, ``Neural networks and physical systems with emergent collective computational abilities.'' \emph{Proceedings of the national academy of sciences}, vol.~79, no.~8, pp. 2554--2558, 1982.

\bibitem{kohonen1982self}
T.~Kohonen, ``Self-organized formation of topologically correct feature maps,'' \emph{Biological cybernetics}, vol.~43, no.~1, pp. 59--69, 1982.

\bibitem{mead2020we}
C.~Mead, ``How we created neuromorphic engineering,'' \emph{Nature Electronics}, vol.~3, no.~7, pp. 434--435, 2020.

\bibitem{mahowald1994silicon}
M.~Mahowald and M.~Mahowald, ``The silicon retina,'' \emph{An Analog VLSI System for Stereoscopic Vision}, pp. 4--65, 1994.

\bibitem{mahowald1991silicon}
M.~Mahowald and R.~Douglas, ``A silicon neuron,'' \emph{Nature}, vol. 354, no. 6354, pp. 515--518, 1991.

\bibitem{schemmel2010wafer}
J.~Schemmel, D.~Br{\"u}derle, A.~Gr{\"u}bl, M.~Hock, K.~Meier, and S.~Millner, ``A wafer-scale neuromorphic hardware system for large-scale neural modeling,'' in \emph{2010 ieee international symposium on circuits and systems (iscas)}.\hskip 1em plus 0.5em minus 0.4em\relax IEEE, 2010, pp. 1947--1950.

\bibitem{merolla2014million}
P.~A. Merolla, J.~V. Arthur, R.~Alvarez-Icaza, A.~S. Cassidy, J.~Sawada, F.~Akopyan, B.~L. Jackson, N.~Imam, C.~Guo, Y.~Nakamura \emph{et~al.}, ``A million spiking-neuron integrated circuit with a scalable communication network and interface,'' \emph{Science}, vol. 345, no. 6197, pp. 668--673, 2014.

\bibitem{benjamin2014neurogrid}
B.~V. Benjamin, P.~Gao, E.~McQuinn, S.~Choudhary, A.~R. Chandrasekaran, J.-M. Bussat, R.~Alvarez-Icaza, J.~V. Arthur, P.~A. Merolla, and K.~Boahen, ``Neurogrid: A mixed-analog-digital multichip system for large-scale neural simulations,'' \emph{Proceedings of the IEEE}, vol. 102, no.~5, pp. 699--716, 2014.

\bibitem{furber2014spinnaker}
S.~B. Furber, F.~Galluppi, S.~Temple, and L.~A. Plana, ``The spinnaker project,'' \emph{Proceedings of the IEEE}, vol. 102, no.~5, pp. 652--665, 2014.

\bibitem{davies2018loihi}
M.~Davies, N.~Srinivasa, T.-H. Lin, G.~Chinya, Y.~Cao, S.~H. Choday, G.~Dimou, P.~Joshi, N.~Imam, S.~Jain \emph{et~al.}, ``Loihi: A neuromorphic manycore processor with on-chip learning,'' \emph{Ieee Micro}, vol.~38, no.~1, pp. 82--99, 2018.

\bibitem{mead1989analog}
C.~Mead and M.~Ismail, \emph{Analog VLSI implementation of neural systems}.\hskip 1em plus 0.5em minus 0.4em\relax Springer Science \& Business Media, 1989, vol.~80.

\bibitem{rose2021system}
G.~S. Rose, M.~S.~A. Shawkat, A.~Z. Foshie, J.~J. Murray, and M.~M. Adnan, ``A system design perspective on neuromorphic computer processors,'' \emph{Neuromorphic Computing and Engineering}, vol.~1, no.~2, p. 022001, 2021.

\bibitem{Najem2018MemristiveMimics}
J.~S. Najem, G.~J. Taylor, R.~J. Weiss, M.~S. Hasan, G.~Rose, C.~D. Schuman, A.~Belianinov, C.~P. Collier, and S.~A. Sarles, ``{Memristive Ion Channel-Doped Biomembranes as Synaptic Mimics},'' \emph{ACS Nano}, vol.~12, no.~5, pp. 4702--4711, 5 2018.

\bibitem{krnjevic1974chemical}
K.~Krnjevi{\'c}, ``Chemical nature of synaptic transmission in vertebrates,'' \emph{Physiological Reviews}, vol.~54, no.~2, pp. 418--540, 1974.

\bibitem{haas2011activity}
J.~S. Haas, B.~Zavala, and C.~E. Landisman, ``Activity-dependent long-term depression of electrical synapses,'' \emph{Science}, vol. 334, no. 6054, pp. 389--393, 2011.

\bibitem{haas2016activity}
J.~S. Haas, C.~M. Greenwald, and A.~E. Pereda, ``Activity-dependent plasticity of electrical synapses: increasing evidence for its presence and functional roles in the mammalian brain,'' \emph{BMC cell biology}, vol.~17, no.~1, pp. 51--57, 2016.

\bibitem{maass1997networks}
W.~Maass, ``Networks of spiking neurons: the third generation of neural network models,'' \emph{Neural networks}, vol.~10, no.~9, pp. 1659--1671, 1997.

\bibitem{izhikevich2003simple}
E.~M. Izhikevich, ``Simple model of spiking neurons,'' \emph{IEEE Transactions on neural networks}, vol.~14, no.~6, pp. 1569--1572, 2003.

\bibitem{frenkel2021bottom}
C.~Frenkel, D.~Bol, and G.~Indiveri, ``Bottom-up and top-down neural processing systems design: Neuromorphic intelligence as the convergence of natural and artificial intelligence,'' \emph{arXiv preprint arXiv:2106.01288}, 2021.

\bibitem{camunas2019neuromorphic}
L.~A. Camu{\~n}as-Mesa, B.~Linares-Barranco, and T.~Serrano-Gotarredona, ``Neuromorphic spiking neural networks and their memristor-cmos hardware implementations,'' \emph{Materials}, vol.~12, no.~17, p. 2745, 2019.

\bibitem{thorpe1996speed}
S.~Thorpe, D.~Fize, and C.~Marlot, ``Speed of processing in the human visual system,'' \emph{nature}, vol. 381, no. 6582, pp. 520--522, 1996.

\bibitem{rolls1994processing}
E.~T. Rolls and M.~J. Tovee, ``Processing speed in the cerebral cortex and the neurophysiology of visual masking,'' \emph{Proceedings of the Royal Society of London. Series B: Biological Sciences}, vol. 257, no. 1348, pp. 9--15, 1994.

\bibitem{hendy2022review}
H.~Hendy and C.~Merkel, ``Review of spike-based neuromorphic computing for brain-inspired vision: biology, algorithms, and hardware,'' \emph{Journal of Electronic Imaging}, vol.~31, no.~1, pp. 010\,901--010\,901, 2022.

\bibitem{roy2019towards}
K.~Roy, A.~Jaiswal, and P.~Panda, ``Towards spike-based machine intelligence with neuromorphic computing,'' \emph{Nature}, vol. 575, no. 7784, pp. 607--617, 2019.

\bibitem{brette2005adaptive}
R.~Brette and W.~Gerstner, ``Adaptive exponential integrate-and-fire model as an effective description of neuronal activity,'' \emph{Journal of neurophysiology}, vol.~94, no.~5, pp. 3637--3642, 2005.

\bibitem{gerstner1993spikes}
W.~Gerstner, R.~Ritz, and J.~L. Van~Hemmen, ``Why spikes? hebbian learning and retrieval of time-resolved excitation patterns,'' \emph{Biological cybernetics}, vol.~69, no. 5-6, pp. 503--515, 1993.

\bibitem{gerstner2002spiking}
W.~Gerstner and W.~M. Kistler, \emph{Spiking neuron models: Single neurons, populations, plasticity}.\hskip 1em plus 0.5em minus 0.4em\relax Cambridge university press, 2002.

\bibitem{burroni2017energetic}
J.~Burroni, P.~Taylor, C.~Corey, T.~Vachnadze, and H.~T. Siegelmann, ``Energetic constraints produce self-sustained oscillatory dynamics in neuronal networks,'' \emph{Frontiers in neuroscience}, vol.~11, p.~80, 2017.

\bibitem{izhikevich2004model}
E.~M. Izhikevich, ``Which model to use for cortical spiking neurons?'' \emph{IEEE transactions on neural networks}, vol.~15, no.~5, pp. 1063--1070, 2004.

\bibitem{indiveri2006vlsi}
G.~Indiveri, E.~Chicca, and R.~Douglas, ``A vlsi array of low-power spiking neurons and bistable synapses with spike-timing dependent plasticity,'' \emph{IEEE transactions on neural networks}, vol.~17, no.~1, pp. 211--221, 2006.

\bibitem{koickal2007analog}
T.~J. Koickal, A.~Hamilton, S.~L. Tan, J.~A. Covington, J.~W. Gardner, and T.~C. Pearce, ``Analog vlsi circuit implementation of an adaptive neuromorphic olfaction chip,'' \emph{IEEE Transactions on Circuits and Systems I: Regular Papers}, vol.~54, no.~1, pp. 60--73, 2007.

\bibitem{ebong2011cmos}
I.~E. Ebong and P.~Mazumder, ``Cmos and memristor-based neural network design for position detection,'' \emph{Proceedings of the IEEE}, vol. 100, no.~6, pp. 2050--2060, 2011.

\bibitem{tovar2008analog}
G.~M. Tovar, E.~S. Fukuda, T.~Asai, T.~Hirose, and Y.~Amemiya, ``Analog cmos circuits implementing neural segmentation model based on symmetric stdp learning,'' in \emph{Neural Information Processing: 14th International Conference, ICONIP 2007, Kitakyushu, Japan, November 13-16, 2007, Revised Selected Papers, Part II 14}.\hskip 1em plus 0.5em minus 0.4em\relax Springer, 2008, pp. 117--126.

\bibitem{indiveri2003low}
G.~Indiveri, ``A low-power adaptive integrate-and-fire neuron circuit,'' in \emph{Proceedings of the 2003 International Symposium on Circuits and Systems, 2003. ISCAS'03.}, vol.~4.\hskip 1em plus 0.5em minus 0.4em\relax IEEE, 2003, pp. IV--IV.

\bibitem{indiveri2010spike}
G.~Indiveri, F.~Stefanini, and E.~Chicca, ``Spike-based learning with a generalized integrate and fire silicon neuron,'' in \emph{Proceedings of 2010 IEEE International Symposium on Circuits and Systems}.\hskip 1em plus 0.5em minus 0.4em\relax IEEE, 2010, pp. 1951--1954.

\bibitem{wu2014energy}
X.~Wu, V.~Saxena, and K.~A. Campbell, ``Energy-efficient stdp-based learning circuits with memristor synapses,'' in \emph{Machine Intelligence and Bio-inspired Computation: Theory and Applications VIII}, vol. 9119.\hskip 1em plus 0.5em minus 0.4em\relax SPIE, 2014, pp. 33--39.

\bibitem{liu2014event}
S.-C. Liu, T.~Delbruck, G.~Indiveri, A.~Whatley, and R.~Douglas, \emph{Event-based neuromorphic systems}.\hskip 1em plus 0.5em minus 0.4em\relax John Wiley \& Sons, 2014.

\bibitem{skrbek1999fast}
M.~Skrbek, ``Fast neural network implementation,'' \emph{Neural Network World}, vol.~9, no.~5, pp. 375--391, 1999.

\bibitem{hikawa2003digital}
H.~Hikawa, ``A digital hardware pulse-mode neuron with piecewise linear activation function,'' \emph{IEEE Transactions on Neural Networks}, vol.~14, no.~5, pp. 1028--1037, 2003.

\bibitem{muthuramalingam2008neural}
A.~Muthuramalingam, S.~Himavathi, and E.~Srinivasan, ``Neural network implementation using fpga: issues and application,'' \emph{International Journal of Electrical and Computer Engineering}, vol.~2, no.~12, pp. 2802--2808, 2008.

\bibitem{sayyaparaju2020device}
S.~Sayyaparaju, M.~M. Adnan, S.~Amer, and G.~S. Rose, ``Device-aware circuit design for robust memristive neuromorphic systems with stdp-based learning,'' \emph{ACM Journal on Emerging Technologies in Computing Systems (JETC)}, vol.~16, no.~3, pp. 1--25, 2020.

\bibitem{zhu2020comprehensive}
J.~Zhu, T.~Zhang, Y.~Yang, and R.~Huang, ``A comprehensive review on emerging artificial neuromorphic devices,'' \emph{Applied Physics Reviews}, vol.~7, no.~1, p. 011312, 2020.

\bibitem{islam2019device}
R.~Islam, H.~Li, P.-Y. Chen, W.~Wan, H.-Y. Chen, B.~Gao, H.~Wu, S.~Yu, K.~Saraswat, and H.~P. Wong, ``Device and materials requirements for neuromorphic computing,'' \emph{Journal of Physics D: Applied Physics}, vol.~52, no.~11, p. 113001, 2019.

\bibitem{ielmini2019emerging}
D.~Ielmini and S.~Ambrogio, ``Emerging neuromorphic devices,'' \emph{Nanotechnology}, vol.~31, no.~9, p. 092001, 2019.

\bibitem{wang2020resistive}
Z.~Wang, H.~Wu, G.~W. Burr, C.~S. Hwang, K.~L. Wang, Q.~Xia, and J.~J. Yang, ``Resistive switching materials for information processing,'' \emph{Nature Reviews Materials}, vol.~5, no.~3, pp. 173--195, 2020.

\bibitem{strukov2008missing}
D.~B. Strukov, G.~S. Snider, D.~R. Stewart, and R.~S. Williams, ``The missing memristor found,'' \emph{nature}, vol. 453, no. 7191, pp. 80--83, 2008.

\bibitem{kumar2022dynamical}
S.~Kumar, X.~Wang, J.~P. Strachan, Y.~Yang, and W.~D. Lu, ``Dynamical memristors for higher-complexity neuromorphic computing,'' \emph{Nature Reviews Materials}, vol.~7, no.~7, pp. 575--591, 2022.

\bibitem{nawrocki2016mini}
R.~A. Nawrocki, R.~M. Voyles, and S.~E. Shaheen, ``A mini review of neuromorphic architectures and implementations,'' \emph{IEEE Transactions on Electron Devices}, vol.~63, no.~10, pp. 3819--3829, 2016.

\bibitem{he2021recent}
Y.~He, L.~Zhu, Y.~Zhu, C.~Chen, S.~Jiang, R.~Liu, Y.~Shi, and Q.~Wan, ``Recent progress on emerging transistor-based neuromorphic devices,'' \emph{Advanced Intelligent Systems}, vol.~3, no.~7, p. 2000210, 2021.

\bibitem{chua1971memristor}
L.~Chua, ``Memristor-the missing circuit element,'' \emph{IEEE Transactions on circuit theory}, vol.~18, no.~5, pp. 507--519, 1971.

\bibitem{pickett2013scalable}
M.~D. Pickett, G.~Medeiros-Ribeiro, and R.~S. Williams, ``A scalable neuristor built with mott memristors,'' \emph{Nature materials}, vol.~12, no.~2, pp. 114--117, 2013.

\bibitem{zhu2019ovonic}
M.~Zhu, K.~Ren, and Z.~Song, ``Ovonic threshold switching selectors for three-dimensional stackable phase-change memory,'' \emph{MRS Bulletin}, vol.~44, no.~9, pp. 715--720, 2019.

\bibitem{wang2017memristors}
Z.~Wang, S.~Joshi, S.~E. Savel’ev, H.~Jiang, R.~Midya, P.~Lin, M.~Hu, N.~Ge, J.~P. Strachan, Z.~Li \emph{et~al.}, ``Memristors with diffusive dynamics as synaptic emulators for neuromorphic computing,'' \emph{Nature materials}, vol.~16, no.~1, pp. 101--108, 2017.

\bibitem{kumar2016direct}
S.~Kumar, C.~E. Graves, J.~P. Strachan, E.~M. Grafals, A.~L.~D. Kilcoyne, T.~Tyliszczak, J.~N. Weker, Y.~Nishi, and R.~S. Williams, ``Direct observation of localized radial oxygen migration in functioning tantalum oxide memristors,'' \emph{Advanced Materials}, vol.~28, no.~14, pp. 2772--2776, 2016.

\bibitem{pershin2008spin}
Y.~V. Pershin and M.~Di~Ventra, ``Spin memristive systems: Spin memory effects in semiconductor spintronics,'' \emph{Physical Review B}, vol.~78, no.~11, p. 113309, 2008.

\bibitem{manipatruni2018beyond}
S.~Manipatruni, D.~E. Nikonov, and I.~A. Young, ``Beyond cmos computing with spin and polarization,'' \emph{Nature Physics}, vol.~14, no.~4, pp. 338--343, 2018.

\bibitem{endoh2016overview}
T.~Endoh, H.~Koike, S.~Ikeda, T.~Hanyu, and H.~Ohno, ``An overview of nonvolatile emerging memories—spintronics for working memories,'' \emph{IEEE journal on emerging and selected topics in circuits and systems}, vol.~6, no.~2, pp. 109--119, 2016.

\bibitem{raoux2014phase}
S.~Raoux, F.~Xiong, M.~Wuttig, and E.~Pop, ``Phase change materials and phase change memory,'' \emph{MRS bulletin}, vol.~39, no.~8, pp. 703--710, 2014.

\bibitem{wang2019overview}
H.~Wang and X.~Yan, ``Overview of resistive random access memory (rram): Materials, filament mechanisms, performance optimization, and prospects,'' \emph{physica status solidi (RRL)--Rapid Research Letters}, vol.~13, no.~9, p. 1900073, 2019.

\bibitem{hong2018oxide}
X.~Hong, D.~J. Loy, P.~A. Dananjaya, F.~Tan, C.~Ng, and W.~Lew, ``Oxide-based rram materials for neuromorphic computing,'' \emph{Journal of materials science}, vol.~53, pp. 8720--8746, 2018.

\bibitem{rehman2020decade}
M.~M. Rehman, H.~M. M.~U. Rehman, J.~Z. Gul, W.~Y. Kim, K.~S. Karimov, and N.~Ahmed, ``Decade of 2d-materials-based rram devices: a review,'' \emph{Science and technology of advanced materials}, vol.~21, no.~1, pp. 147--186, 2020.

\bibitem{feng20202d}
X.~Feng, X.~Liu, and K.-W. Ang, ``2d photonic memristor beyond graphene: progress and prospects,'' \emph{Nanophotonics}, vol.~9, no.~7, pp. 1579--1599, 2020.

\bibitem{wang2019mott}
Y.~Wang, K.-M. Kang, M.~Kim, H.-S. Lee, R.~Waser, D.~Wouters, R.~Dittmann, J.~J. Yang, and H.-H. Park, ``Mott-transition-based rram,'' \emph{Materials today}, vol.~28, pp. 63--80, 2019.

\bibitem{goswami2020organic}
S.~Goswami, S.~Goswami, and T.~Venkatesan, ``An organic approach to low energy memory and brain inspired electronics,'' \emph{Applied Physics Reviews}, vol.~7, no.~2, p. 021303, 2020.

\bibitem{ahn2018carbon}
E.~C. Ahn, H.-S.~P. Wong, and E.~Pop, ``Carbon nanomaterials for non-volatile memories,'' \emph{Nature Reviews Materials}, vol.~3, no.~3, pp. 1--15, 2018.

\bibitem{kim2015experimental}
S.~Kim, C.~Du, P.~Sheridan, W.~Ma, S.~Choi, and W.~D. Lu, ``Experimental demonstration of a second-order memristor and its ability to biorealistically implement synaptic plasticity,'' \emph{Nano letters}, vol.~15, no.~3, pp. 2203--2211, 2015.

\bibitem{rodriguez2018characterization}
A.~Rodriguez-Fernandez, C.~Cagli, L.~Perniola, E.~Miranda, and J.~Su{\~n}{\'e}, ``Characterization of hfo2-based devices with indication of second order memristor effects,'' \emph{Microelectronic Engineering}, vol. 195, pp. 101--106, 2018.

\bibitem{mikheev2019ferroelectric}
V.~Mikheev, A.~Chouprik, Y.~Lebedinskii, S.~Zarubin, Y.~Matveyev, E.~Kondratyuk, M.~G. Kozodaev, A.~M. Markeev, A.~Zenkevich, and D.~Negrov, ``Ferroelectric second-order memristor,'' \emph{ACS applied materials \& interfaces}, vol.~11, no.~35, pp. 32\,108--32\,114, 2019.

\bibitem{najem2018memristive}
J.~S. Najem, G.~J. Taylor, R.~J. Weiss, M.~S. Hasan, G.~Rose, C.~D. Schuman, A.~Belianinov, C.~P. Collier, and S.~A. Sarles, ``Memristive ion channel-doped biomembranes as synaptic mimics,'' \emph{ACS nano}, vol.~12, no.~5, pp. 4702--4711, 2018.

\bibitem{hasan2018biomimetic}
M.~S. Hasan, C.~D. Schuman, J.~S. Najem, R.~Weiss, N.~D. Skuda, A.~Belianinov, C.~P. Collier, S.~A. Sarles, and G.~S. Rose, ``Biomimetic, soft-material synapse for neuromorphic computing: from device to network,'' in \emph{2018 IEEE 13th Dallas Circuits and Systems Conference (DCAS)}.\hskip 1em plus 0.5em minus 0.4em\relax IEEE, 2018, pp. 1--6.

\bibitem{koner2019memristive}
S.~Koner, J.~S. Najem, M.~S. Hasan, and S.~A. Sarles, ``Memristive plasticity in artificial electrical synapses via geometrically reconfigurable, gramicidin-doped biomembranes,'' \emph{Nanoscale}, vol.~11, no.~40, pp. 18\,640--18\,652, 2019.

\bibitem{alexandrov2011current}
A.~Alexandrov, A.~Bratkovsky, B.~Bridle, S.~Savel’Ev, D.~Strukov, and R.~Stanley~Williams, ``Current-controlled negative differential resistance due to joule heating in tio2,'' \emph{Applied Physics Letters}, vol.~99, no.~20, p. 202104, 2011.

\bibitem{pickett2011coexistence}
M.~D. Pickett, J.~Borghetti, J.~J. Yang, G.~Medeiros-Ribeiro, and R.~S. Williams, ``Coexistence of memristance and negative differential resistance in a nanoscale metal-oxide-metal system,'' \emph{Advanced Materials}, vol.~23, no.~15, pp. 1730--1733, 2011.

\bibitem{kumar2020third}
S.~Kumar, R.~S. Williams, and Z.~Wang, ``Third-order nanocircuit elements for neuromorphic engineering,'' \emph{Nature}, vol. 585, no. 7826, pp. 518--523, 2020.

\bibitem{du2017reservoir}
C.~Du, F.~Cai, M.~A. Zidan, W.~Ma, S.~H. Lee, and W.~D. Lu, ``Reservoir computing using dynamic memristors for temporal information processing,'' \emph{Nature communications}, vol.~8, no.~1, pp. 1--10, 2017.

\bibitem{moon2019temporal}
J.~Moon, W.~Ma, J.~H. Shin, F.~Cai, C.~Du, S.~H. Lee, and W.~D. Lu, ``Temporal data classification and forecasting using a memristor-based reservoir computing system,'' \emph{Nature Electronics}, vol.~2, no.~10, pp. 480--487, 2019.

\bibitem{zhong2022memristor}
Y.~Zhong, J.~Tang, X.~Li, X.~Liang, Z.~Liu, Y.~Li, Y.~Xi, P.~Yao, Z.~Hao, B.~Gao \emph{et~al.}, ``A memristor-based analogue reservoir computing system for real-time and power-efficient signal processing,'' \emph{Nature Electronics}, vol.~5, no.~10, pp. 672--681, 2022.

\bibitem{hossain2023energy}
M.~R. Hossain, A.~S. Mohamed, N.~X. Armendarez, J.~S. Najem, and M.~S. Hasan, ``Energy-efficient memcapacitive physical reservoir computing system for temporal data processing,'' \emph{arXiv preprint arXiv:2305.12025}, 2023.

\bibitem{moradi2017scalable}
S.~Moradi, N.~Qiao, F.~Stefanini, and G.~Indiveri, ``A scalable multicore architecture with heterogeneous memory structures for dynamic neuromorphic asynchronous processors (dynaps),'' \emph{IEEE transactions on biomedical circuits and systems}, vol.~12, no.~1, pp. 106--122, 2017.

\bibitem{thakur2018large}
C.~S. Thakur, J.~L. Molin, G.~Cauwenberghs, G.~Indiveri, K.~Kumar, N.~Qiao, J.~Schemmel, R.~Wang, E.~Chicca, J.~Olson~Hasler \emph{et~al.}, ``Large-scale neuromorphic spiking array processors: A quest to mimic the brain,'' \emph{Frontiers in neuroscience}, vol.~12, p. 891, 2018.

\bibitem{amunts2016human}
K.~Amunts, C.~Ebell, J.~Muller, M.~Telefont, A.~Knoll, and T.~Lippert, ``The human brain project: creating a european research infrastructure to decode the human brain,'' \emph{Neuron}, vol.~92, no.~3, pp. 574--581, 2016.

\bibitem{dean2014dynamic}
M.~E. Dean, C.~D. Schuman, and J.~D. Birdwell, ``Dynamic adaptive neural network array,'' in \emph{Unconventional Computation and Natural Computation: 13th International Conference, UCNC 2014, London, ON, Canada, July 14-18, 2014, Proceedings 13}.\hskip 1em plus 0.5em minus 0.4em\relax Springer, 2014, pp. 129--141.

\bibitem{mitchell2018danna}
J.~P. Mitchell, M.~E. Dean, G.~R. Bruer, J.~S. Plank, and G.~S. Rose, ``Danna 2: Dynamic adaptive neural network arrays,'' in \emph{Proceedings of the International Conference on Neuromorphic Systems}, 2018, pp. 1--6.

\bibitem{chakma2017memristive}
G.~Chakma, M.~M. Adnan, A.~R. Wyer, R.~Weiss, C.~D. Schuman, and G.~S. Rose, ``Memristive mixed-signal neuromorphic systems: Energy-efficient learning at the circuit-level,'' \emph{IEEE Journal on Emerging and Selected Topics in Circuits and Systems}, vol.~8, no.~1, pp. 125--136, 2017.

\bibitem{foshie2021multi}
A.~Z. Foshie, N.~N. Chakraborty, J.~J. Murray, T.~J. Fowler, M.~S.~A. Shawkat, and G.~S. Rose, ``A multi-context neural core design for reconfigurable neuromorphic arrays,'' in \emph{2021 IEEE Computer Society Annual Symposium on VLSI (ISVLSI)}.\hskip 1em plus 0.5em minus 0.4em\relax IEEE, 2021, pp. 67--72.

\bibitem{hasler2019large}
J.~Hasler, ``Large-scale field-programmable analog arrays,'' \emph{Proceedings of the IEEE}, vol. 108, no.~8, pp. 1283--1302, 2019.

\bibitem{diehl2015fast}
P.~U. Diehl, D.~Neil, J.~Binas, M.~Cook, S.-C. Liu, and M.~Pfeiffer, ``Fast-classifying, high-accuracy spiking deep networks through weight and threshold balancing,'' in \emph{2015 International joint conference on neural networks (IJCNN)}.\hskip 1em plus 0.5em minus 0.4em\relax ieee, 2015, pp. 1--8.

\bibitem{diehl2016conversion}
P.~U. Diehl, G.~Zarrella, A.~Cassidy, B.~U. Pedroni, and E.~Neftci, ``Conversion of artificial recurrent neural networks to spiking neural networks for low-power neuromorphic hardware,'' in \emph{2016 IEEE International Conference on Rebooting Computing (ICRC)}.\hskip 1em plus 0.5em minus 0.4em\relax IEEE, 2016, pp. 1--8.

\bibitem{hunsberger2016training}
E.~Hunsberger and C.~Eliasmith, ``Training spiking deep networks for neuromorphic hardware,'' \emph{arXiv preprint arXiv:1611.05141}, 2016.

\bibitem{sengupta2019going}
A.~Sengupta, Y.~Ye, R.~Wang, C.~Liu, and K.~Roy, ``Going deeper in spiking neural networks: Vgg and residual architectures,'' \emph{Frontiers in neuroscience}, vol.~13, p.~95, 2019.

\bibitem{rueckauer2017conversion}
B.~Rueckauer, I.-A. Lungu, Y.~Hu, M.~Pfeiffer, and S.-C. Liu, ``Conversion of continuous-valued deep networks to efficient event-driven networks for image classification,'' \emph{Frontiers in neuroscience}, vol.~11, p. 682, 2017.

\bibitem{neil2016learning}
D.~Neil, M.~Pfeiffer, and S.-C. Liu, ``Learning to be efficient: Algorithms for training low-latency, low-compute deep spiking neural networks,'' in \emph{Proceedings of the 31st annual ACM symposium on applied computing}, 2016, pp. 293--298.

\bibitem{esser2015backpropagation}
S.~K. Esser, R.~Appuswamy, P.~Merolla, J.~V. Arthur, and D.~S. Modha, ``Backpropagation for energy-efficient neuromorphic computing,'' \emph{Advances in neural information processing systems}, vol.~28, 2015.

\bibitem{cao2015spiking}
Y.~Cao, Y.~Chen, and D.~Khosla, ``Spiking deep convolutional neural networks for energy-efficient object recognition,'' \emph{International Journal of Computer Vision}, vol. 113, pp. 54--66, 2015.

\bibitem{bohte2000spikeprop}
S.~M. Bohte, J.~N. Kok, and J.~A. La~Poutr{\'e}, ``Spikeprop: backpropagation for networks of spiking neurons.'' in \emph{ESANN}, vol.~48.\hskip 1em plus 0.5em minus 0.4em\relax Bruges, 2000, pp. 419--424.

\bibitem{severa2019training}
W.~Severa, C.~M. Vineyard, R.~Dellana, S.~J. Verzi, and J.~B. Aimone, ``Training deep neural networks for binary communication with the whetstone method,'' \emph{Nature Machine Intelligence}, vol.~1, no.~2, pp. 86--94, 2019.

\bibitem{wu2018spatio}
Y.~Wu, L.~Deng, G.~Li, J.~Zhu, and L.~Shi, ``Spatio-temporal backpropagation for training high-performance spiking neural networks,'' \emph{Frontiers in neuroscience}, vol.~12, p. 331, 2018.

\bibitem{neftci2019surrogate}
E.~O. Neftci, H.~Mostafa, and F.~Zenke, ``Surrogate gradient learning in spiking neural networks: Bringing the power of gradient-based optimization to spiking neural networks,'' \emph{IEEE Signal Processing Magazine}, vol.~36, no.~6, pp. 51--63, 2019.

\bibitem{lee2016training}
J.~H. Lee, T.~Delbruck, and M.~Pfeiffer, ``Training deep spiking neural networks using backpropagation,'' \emph{Frontiers in neuroscience}, vol.~10, p. 508, 2016.

\bibitem{bellec2018long}
G.~Bellec, D.~Salaj, A.~Subramoney, R.~Legenstein, and W.~Maass, ``Long short-term memory and learning-to-learn in networks of spiking neurons,'' \emph{Advances in neural information processing systems}, vol.~31, 2018.

\bibitem{meng2022training}
Q.~Meng, M.~Xiao, S.~Yan, Y.~Wang, Z.~Lin, and Z.-Q. Luo, ``Training high-performance low-latency spiking neural networks by differentiation on spike representation,'' in \emph{Proceedings of the IEEE/CVF Conference on Computer Vision and Pattern Recognition}, 2022, pp. 12\,444--12\,453.

\bibitem{tavanaei2019deep}
A.~Tavanaei, M.~Ghodrati, S.~R. Kheradpisheh, T.~Masquelier, and A.~Maida, ``Deep learning in spiking neural networks,'' \emph{Neural networks}, vol. 111, pp. 47--63, 2019.

\bibitem{caporale2008spike}
N.~Caporale and Y.~Dan, ``Spike timing--dependent plasticity: a hebbian learning rule,'' \emph{Annu. Rev. Neurosci.}, vol.~31, pp. 25--46, 2008.

\bibitem{schuman2017survey}
C.~D. Schuman, T.~E. Potok, R.~M. Patton, J.~D. Birdwell, M.~E. Dean, G.~S. Rose, and J.~S. Plank, ``A survey of neuromorphic computing and neural networks in hardware,'' \emph{arXiv preprint arXiv:1705.06963}, 2017.

\bibitem{diehl2015unsupervised}
P.~U. Diehl and M.~Cook, ``Unsupervised learning of digit recognition using spike-timing-dependent plasticity,'' \emph{Frontiers in computational neuroscience}, vol.~9, p.~99, 2015.

\bibitem{tavanaei2018training}
A.~Tavanaei, Z.~Kirby, and A.~S. Maida, ``Training spiking convnets by stdp and gradient descent,'' in \emph{2018 International Joint Conference on Neural Networks (IJCNN)}.\hskip 1em plus 0.5em minus 0.4em\relax IEEE, 2018, pp. 1--8.

\bibitem{kheradpisheh2018stdp}
S.~R. Kheradpisheh, M.~Ganjtabesh, S.~J. Thorpe, and T.~Masquelier, ``Stdp-based spiking deep convolutional neural networks for object recognition,'' \emph{Neural Networks}, vol.~99, pp. 56--67, 2018.

\bibitem{iranmehr2020ils}
E.~Iranmehr, S.~B. Shouraki, and M.~M. Faraji, ``Ils-based reservoir computing for handwritten digits recognition,'' in \emph{2020 8th Iranian joint congress on fuzzy and intelligent systems (CFIS)}.\hskip 1em plus 0.5em minus 0.4em\relax IEEE, 2020, pp. 1--6.

\bibitem{reynolds2019intelligent}
J.~J. Reynolds, J.~S. Plank, and C.~D. Schuman, ``Intelligent reservoir generation for liquid state machines using evolutionary optimization,'' in \emph{2019 International Joint Conference on Neural Networks (IJCNN)}.\hskip 1em plus 0.5em minus 0.4em\relax IEEE, 2019, pp. 1--8.

\bibitem{tang2022evolutionary}
C.~Tang, J.~Ji, Q.~Lin, and Y.~Zhou, ``Evolutionary neural architecture design of liquid state machine for image classification,'' in \emph{ICASSP 2022-2022 IEEE International Conference on Acoustics, Speech and Signal Processing (ICASSP)}.\hskip 1em plus 0.5em minus 0.4em\relax IEEE, 2022, pp. 91--95.

\bibitem{wijesinghe2019analysis}
P.~Wijesinghe, G.~Srinivasan, P.~Panda, and K.~Roy, ``Analysis of liquid ensembles for enhancing the performance and accuracy of liquid state machines,'' \emph{Frontiers in neuroscience}, vol.~13, p. 504, 2019.

\bibitem{soriano2014delay}
M.~C. Soriano, S.~Ort{\'\i}n, L.~Keuninckx, L.~Appeltant, J.~Danckaert, L.~Pesquera, and G.~Van~der Sande, ``Delay-based reservoir computing: noise effects in a combined analog and digital implementation,'' \emph{IEEE transactions on neural networks and learning systems}, vol.~26, no.~2, pp. 388--393, 2014.

\bibitem{schrauwen2008compact}
B.~Schrauwen, M.~D’Haene, D.~Verstraeten, and J.~Van~Campenhout, ``Compact hardware liquid state machines on fpga for real-time speech recognition,'' \emph{Neural networks}, vol.~21, no. 2-3, pp. 511--523, 2008.

\bibitem{zhang2015digital}
Y.~Zhang, P.~Li, Y.~Jin, and Y.~Choe, ``A digital liquid state machine with biologically inspired learning and its application to speech recognition,'' \emph{IEEE transactions on neural networks and learning systems}, vol.~26, no.~11, pp. 2635--2649, 2015.

\bibitem{soures2017robustness}
N.~Soures, L.~Hays, and D.~Kudithipudi, ``Robustness of a memristor based liquid state machine,'' in \emph{2017 international joint conference on neural networks (ijcnn)}.\hskip 1em plus 0.5em minus 0.4em\relax IEEE, 2017, pp. 2414--2420.

\bibitem{van2017advances}
G.~Van~der Sande, D.~Brunner, and M.~C. Soriano, ``Advances in photonic reservoir computing,'' \emph{Nanophotonics}, vol.~6, no.~3, pp. 561--576, 2017.

\bibitem{brunner2018tutorial}
D.~Brunner, B.~Penkovsky, B.~A. Marquez, M.~Jacquot, I.~Fischer, and L.~Larger, ``Tutorial: Photonic neural networks in delay systems,'' \emph{Journal of Applied Physics}, vol. 124, no.~15, p. 152004, 2018.

\bibitem{hauser2011towards}
H.~Hauser, A.~J. Ijspeert, R.~M. F{\"u}chslin, R.~Pfeifer, and W.~Maass, ``Towards a theoretical foundation for morphological computation with compliant bodies,'' \emph{Biological cybernetics}, vol. 105, no.~5, pp. 355--370, 2011.

\bibitem{caluwaerts2014design}
K.~Caluwaerts, J.~Despraz, A.~I{\c{s}}{\c{c}}en, A.~P. Sabelhaus, J.~Bruce, B.~Schrauwen, and V.~SunSpiral, ``Design and control of compliant tensegrity robots through simulation and hardware validation,'' \emph{Journal of the royal society interface}, vol.~11, no.~98, p. 20140520, 2014.

\bibitem{dranias2013short}
M.~R. Dranias, H.~Ju, E.~Rajaram, and A.~M. VanDongen, ``Short-term memory in networks of dissociated cortical neurons,'' \emph{Journal of Neuroscience}, vol.~33, no.~5, pp. 1940--1953, 2013.

\bibitem{sussillo2009generating}
D.~Sussillo and L.~F. Abbott, ``Generating coherent patterns of activity from chaotic neural networks,'' \emph{Neuron}, vol.~63, no.~4, pp. 544--557, 2009.

\bibitem{jones2007there}
B.~Jones, D.~Stekel, J.~Rowe, and C.~Fernando, ``Is there a liquid state machine in the bacterium escherichia coli?'' in \emph{2007 IEEE Symposium on Artificial Life}.\hskip 1em plus 0.5em minus 0.4em\relax Ieee, 2007, pp. 187--191.

\bibitem{fujii2017harnessing}
K.~Fujii and K.~Nakajima, ``Harnessing disordered-ensemble quantum dynamics for machine learning,'' \emph{Physical Review Applied}, vol.~8, no.~2, p. 024030, 2017.

\bibitem{torrejon2017neuromorphic}
J.~Torrejon, M.~Riou, F.~A. Araujo, S.~Tsunegi, G.~Khalsa, D.~Querlioz, P.~Bortolotti, V.~Cros, K.~Yakushiji, A.~Fukushima \emph{et~al.}, ``Neuromorphic computing with nanoscale spintronic oscillators,'' \emph{Nature}, vol. 547, no. 7664, pp. 428--431, 2017.

\bibitem{bourianoff2018potential}
G.~Bourianoff, D.~Pinna, M.~Sitte, and K.~Everschor-Sitte, ``Potential implementation of reservoir computing models based on magnetic skyrmions,'' \emph{Aip Advances}, vol.~8, no.~5, p. 055602, 2018.

\bibitem{tanaka2019recent}
G.~Tanaka, T.~Yamane, J.~B. H{\'e}roux, R.~Nakane, N.~Kanazawa, S.~Takeda, H.~Numata, D.~Nakano, and A.~Hirose, ``Recent advances in physical reservoir computing: A review,'' \emph{Neural Networks}, vol. 115, pp. 100--123, 2019.

\bibitem{maass2004fading}
W.~Maass, T.~Natschl{\"a}ger, and H.~Markram, ``Fading memory and kernel properties of generic cortical microcircuit models,'' \emph{Journal of Physiology-Paris}, vol.~98, no. 4-6, pp. 315--330, 2004.

\bibitem{yildiz2012re}
I.~B. Yildiz, H.~Jaeger, and S.~J. Kiebel, ``Re-visiting the echo state property,'' \emph{Neural networks}, vol.~35, pp. 1--9, 2012.

\bibitem{schuman2020evolutionary}
C.~D. Schuman, J.~P. Mitchell, R.~M. Patton, T.~E. Potok, and J.~S. Plank, ``Evolutionary optimization for neuromorphic systems,'' in \emph{Proceedings of the Neuro-inspired Computational Elements Workshop}, 2020, pp. 1--9.

\bibitem{schuman2016evolutionary}
C.~D. Schuman, J.~S. Plank, A.~Disney, and J.~Reynolds, ``An evolutionary optimization framework for neural networks and neuromorphic architectures,'' in \emph{2016 International Joint Conference on Neural Networks (IJCNN)}.\hskip 1em plus 0.5em minus 0.4em\relax IEEE, 2016, pp. 145--154.

\bibitem{schuman2022evolutionary}
C.~Schuman, R.~Patton, S.~Kulkarni, M.~Parsa, C.~Stahl, N.~Q. Haas, J.~P. Mitchell, S.~Snyder, A.~Nagle, A.~Shanafield \emph{et~al.}, ``Evolutionary vs imitation learning for neuromorphic control at the edge,'' \emph{Neuromorphic Computing and Engineering}, vol.~2, no.~1, p. 014002, 2022.

\bibitem{liu2021survey}
Y.~Liu, Y.~Sun, B.~Xue, M.~Zhang, G.~G. Yen, and K.~C. Tan, ``A survey on evolutionary neural architecture search,'' \emph{IEEE transactions on neural networks and learning systems}, 2021.

\bibitem{mostafa2017fast}
H.~Mostafa, B.~U. Pedroni, S.~Sheik, and G.~Cauwenberghs, ``Fast classification using sparsely active spiking networks,'' in \emph{2017 IEEE International Symposium on Circuits and Systems (ISCAS)}.\hskip 1em plus 0.5em minus 0.4em\relax IEEE, 2017, pp. 1--4.

\bibitem{srinivasan2019restocnet}
G.~Srinivasan and K.~Roy, ``Restocnet: Residual stochastic binary convolutional spiking neural network for memory-efficient neuromorphic computing,'' \emph{Frontiers in neuroscience}, vol.~13, p. 189, 2019.

\bibitem{zhang2020efficient}
M.~Zhang, Z.~Gu, N.~Zheng, D.~Ma, and G.~Pan, ``Efficient spiking neural networks with logarithmic temporal coding,'' \emph{IEEE Access}, vol.~8, pp. 98\,156--98\,167, 2020.

\bibitem{zhang2020temporal}
W.~Zhang and P.~Li, ``Temporal spike sequence learning via backpropagation for deep spiking neural networks,'' \emph{Advances in Neural Information Processing Systems}, vol.~33, pp. 12\,022--12\,033, 2020.

\bibitem{rathi2020enabling}
N.~Rathi, G.~Srinivasan, P.~Panda, and K.~Roy, ``Enabling deep spiking neural networks with hybrid conversion and spike timing dependent backpropagation,'' \emph{arXiv preprint arXiv:2005.01807}, 2020.

\bibitem{han2020deep}
B.~Han and K.~Roy, ``Deep spiking neural network: Energy efficiency through time based coding,'' in \emph{Computer Vision--ECCV 2020: 16th European Conference, Glasgow, UK, August 23--28, 2020, Proceedings, Part X}.\hskip 1em plus 0.5em minus 0.4em\relax Springer, 2020, pp. 388--404.

\bibitem{deng2022temporal}
S.~Deng, Y.~Li, S.~Zhang, and S.~Gu, ``Temporal efficient training of spiking neural network via gradient re-weighting,'' \emph{arXiv preprint arXiv:2202.11946}, 2022.

\bibitem{duan2022temporal}
C.~Duan, J.~Ding, S.~Chen, Z.~Yu, and T.~Huang, ``Temporal effective batch normalization in spiking neural networks,'' \emph{Advances in Neural Information Processing Systems}, vol.~35, pp. 34\,377--34\,390, 2022.

\bibitem{han2020rmp}
B.~Han, G.~Srinivasan, and K.~Roy, ``Rmp-snn: Residual membrane potential neuron for enabling deeper high-accuracy and low-latency spiking neural network,'' in \emph{Proceedings of the IEEE/CVF conference on computer vision and pattern recognition}, 2020, pp. 13\,558--13\,567.

\bibitem{li2021differentiable}
Y.~Li, Y.~Guo, S.~Zhang, S.~Deng, Y.~Hai, and S.~Gu, ``Differentiable spike: Rethinking gradient-descent for training spiking neural networks,'' \emph{Advances in Neural Information Processing Systems}, vol.~34, pp. 23\,426--23\,439, 2021.

\bibitem{li2021free}
Y.~Li, S.~Deng, X.~Dong, R.~Gong, and S.~Gu, ``A free lunch from ann: Towards efficient, accurate spiking neural networks calibration,'' in \emph{International Conference on Machine Learning}.\hskip 1em plus 0.5em minus 0.4em\relax PMLR, 2021, pp. 6316--6325.

\bibitem{yao2023attention}
M.~Yao, G.~Zhao, H.~Zhang, Y.~Hu, L.~Deng, Y.~Tian, B.~Xu, and G.~Li, ``Attention spiking neural networks,'' \emph{IEEE Transactions on Pattern Analysis and Machine Intelligence}, 2023.

\bibitem{rathi2023exploring}
N.~Rathi, I.~Chakraborty, A.~Kosta, A.~Sengupta, A.~Ankit, P.~Panda, and K.~Roy, ``Exploring neuromorphic computing based on spiking neural networks: Algorithms to hardware,'' \emph{ACM Computing Surveys}, vol.~55, no.~12, pp. 1--49, 2023.

\bibitem{lecun1998gradient}
Y.~LeCun, L.~Bottou, Y.~Bengio, and P.~Haffner, ``Gradient-based learning applied to document recognition,'' \emph{Proceedings of the IEEE}, vol.~86, no.~11, pp. 2278--2324, 1998.

\bibitem{krizhevsky2009learning}
A.~Krizhevsky, G.~Hinton \emph{et~al.}, ``Learning multiple layers of features from tiny images,'' 2009.

\bibitem{deng2009imagenet}
J.~Deng, W.~Dong, R.~Socher, L.-J. Li, K.~Li, and L.~Fei-Fei, ``Imagenet: A large-scale hierarchical image database,'' in \emph{2009 IEEE conference on computer vision and pattern recognition}.\hskip 1em plus 0.5em minus 0.4em\relax Ieee, 2009, pp. 248--255.

\bibitem{lichtsteiner2008128}
P.~Lichtsteiner, C.~Posch, and T.~Delbruck, ``A 128 $\times$ 128 120 db 15 $mu$s latency asynchronous temporal contrast vision sensor,'' \emph{IEEE journal of solid-state circuits}, vol.~43, no.~2, pp. 566--576, 2008.

\bibitem{4541871}
C.~Posch, D.~Matolin, and R.~Wohlgenannt, ``An asynchronous time-based image sensor,'' pp. 2130--2133, 2008.

\bibitem{gallego2020event}
G.~Gallego, T.~Delbr{\"u}ck, G.~Orchard, C.~Bartolozzi, B.~Taba, A.~Censi, S.~Leutenegger, A.~J. Davison, J.~Conradt, K.~Daniilidis \emph{et~al.}, ``Event-based vision: A survey,'' \emph{IEEE transactions on pattern analysis and machine intelligence}, vol.~44, no.~1, pp. 154--180, 2020.

\bibitem{posch2010qvga}
C.~Posch, D.~Matolin, and R.~Wohlgenannt, ``A qvga 143 db dynamic range frame-free pwm image sensor with lossless pixel-level video compression and time-domain cds,'' \emph{IEEE Journal of Solid-State Circuits}, vol.~46, no.~1, pp. 259--275, 2010.

\bibitem{brandli2014240}
C.~Brandli, R.~Berner, M.~Yang, S.-C. Liu, and T.~Delbruck, ``A 240$\times$ 180 130 db 3 $\mu$s latency global shutter spatiotemporal vision sensor,'' \emph{IEEE Journal of Solid-State Circuits}, vol.~49, no.~10, pp. 2333--2341, 2014.

\bibitem{finateu20201280x720}
T.~Finateu, A.~Niwa, D.~Matolin, K.~Tsuchimoto, A.~Mascheroni, E.~Reynaud, P.~Mostafalu, F.~Brady, L.~Chotard, F.~LeGoff \emph{et~al.}, ``A 1280x720 back-illuminated stacked temporal contrast event-based vision sensor with 4.86 um pixels, 1.066 geps readout, programmable event-rate controller and compressive data-formatting pipeline,'' in \emph{IEEE International Solid-State Circuits Conference}, 2020.

\bibitem{son20174}
B.~Son, Y.~Suh, S.~Kim, H.~Jung, J.-S. Kim, C.~Shin, K.~Park, K.~Lee, J.~Park, J.~Woo \emph{et~al.}, ``4.1 a 640$\times$ 480 dynamic vision sensor with a 9$\mu$m pixel and 300meps address-event representation,'' in \emph{2017 IEEE International Solid-State Circuits Conference (ISSCC)}.\hskip 1em plus 0.5em minus 0.4em\relax IEEE, 2017, pp. 66--67.

\bibitem{suh20201280}
Y.~Suh, S.~Choi, M.~Ito, J.~Kim, Y.~Lee, J.~Seo, H.~Jung, D.-H. Yeo, S.~Namgung, J.~Bong \emph{et~al.}, ``A 1280$\times$ 960 dynamic vision sensor with a 4.95-$\mu$m pixel pitch and motion artifact minimization,'' in \emph{2020 IEEE international symposium on circuits and systems (ISCAS)}.\hskip 1em plus 0.5em minus 0.4em\relax IEEE, 2020, pp. 1--5.

\bibitem{chen2019live}
S.~Chen and M.~Guo, ``Live demonstration: Celex-v: A 1m pixel multi-mode event-based sensor,'' in \emph{2019 IEEE/CVF Conference on Computer Vision and Pattern Recognition Workshops (CVPRW)}.\hskip 1em plus 0.5em minus 0.4em\relax IEEE, 2019, pp. 1682--1683.

\bibitem{insight}
``Insightness event-based sensor modules,'' \emph{Available: http://www.insightness.com/technology/}, 2020.

\bibitem{delbruck2016neuromorophic}
T.~Delbruck, ``Neuromorophic vision sensing and processing,'' in \emph{2016 46Th european solid-state device research conference (ESSDERC)}.\hskip 1em plus 0.5em minus 0.4em\relax IEEE, 2016, pp. 7--14.

\bibitem{liu2019event}
S.-C. Liu, B.~Rueckauer, E.~Ceolini, A.~Huber, and T.~Delbruck, ``Event-driven sensing for efficient perception: Vision and audition algorithms,'' \emph{IEEE Signal Processing Magazine}, vol.~36, no.~6, pp. 29--37, 2019.

\bibitem{delbruck2013robotic}
T.~Delbruck and M.~Lang, ``Robotic goalie with 3 ms reaction time at 4\% cpu load using event-based dynamic vision sensor,'' \emph{Frontiers in neuroscience}, vol.~7, p. 223, 2013.

\bibitem{glover2016event}
A.~Glover and C.~Bartolozzi, ``Event-driven ball detection and gaze fixation in clutter,'' in \emph{2016 IEEE/RSJ International Conference on Intelligent Robots and Systems (IROS)}.\hskip 1em plus 0.5em minus 0.4em\relax IEEE, 2016, pp. 2203--2208.

\bibitem{litzenberger2006estimation}
M.~Litzenberger, B.~Kohn, A.~N. Belbachir, N.~Donath, G.~Gritsch, H.~Garn, C.~Posch, and S.~Schraml, ``Estimation of vehicle speed based on asynchronous data from a silicon retina optical sensor,'' in \emph{2006 IEEE intelligent transportation systems conference}.\hskip 1em plus 0.5em minus 0.4em\relax IEEE, 2006, pp. 653--658.

\bibitem{lee2014real}
J.~H. Lee, T.~Delbruck, M.~Pfeiffer, P.~K. Park, C.-W. Shin, H.~Ryu, and B.~C. Kang, ``Real-time gesture interface based on event-driven processing from stereo silicon retinas,'' \emph{IEEE transactions on neural networks and learning systems}, vol.~25, no.~12, pp. 2250--2263, 2014.

\bibitem{orchard2015hfirst}
G.~Orchard, C.~Meyer, R.~Etienne-Cummings, C.~Posch, N.~Thakor, and R.~Benosman, ``Hfirst: A temporal approach to object recognition,'' \emph{IEEE transactions on pattern analysis and machine intelligence}, vol.~37, no.~10, pp. 2028--2040, 2015.

\bibitem{rogister2011asynchronous}
P.~Rogister, R.~Benosman, S.-H. Ieng, P.~Lichtsteiner, and T.~Delbruck, ``Asynchronous event-based binocular stereo matching,'' \emph{IEEE Transactions on Neural Networks and Learning Systems}, vol.~23, no.~2, pp. 347--353, 2011.

\bibitem{rebecq2018emvs}
H.~Rebecq, G.~Gallego, E.~Mueggler, and D.~Scaramuzza, ``Emvs: Event-based multi-view stereo—3d reconstruction with an event camera in real-time,'' \emph{International Journal of Computer Vision}, vol. 126, no.~12, pp. 1394--1414, 2018.

\bibitem{matsuda2015mc3d}
N.~Matsuda, O.~Cossairt, and M.~Gupta, ``Mc3d: Motion contrast 3d scanning,'' in \emph{2015 IEEE International Conference on Computational Photography (ICCP)}.\hskip 1em plus 0.5em minus 0.4em\relax IEEE, 2015, pp. 1--10.

\bibitem{benosman2013event}
R.~Benosman, C.~Clercq, X.~Lagorce, S.-H. Ieng, and C.~Bartolozzi, ``Event-based visual flow,'' \emph{IEEE transactions on neural networks and learning systems}, vol.~25, no.~2, pp. 407--417, 2013.

\bibitem{Kim2014SimultaneousMA}
\BIBentryALTinterwordspacing
H.~Kim, A.~Handa, R.~B. Benosman, S.-H. Ieng, and A.~J. Davison, ``Simultaneous mosaicing and tracking with an event camera,'' in \emph{British Machine Vision Conference}, 2014. [Online]. Available: \url{https://api.semanticscholar.org/CorpusID:14402590}
\BIBentrySTDinterwordspacing

\bibitem{rebecq2019high}
H.~Rebecq, R.~Ranftl, V.~Koltun, and D.~Scaramuzza, ``High speed and high dynamic range video with an event camera,'' \emph{IEEE transactions on pattern analysis and machine intelligence}, vol.~43, no.~6, pp. 1964--1980, 2019.

\bibitem{kim2016real}
H.~Kim, S.~Leutenegger, and A.~J. Davison, ``Real-time 3d reconstruction and 6-dof tracking with an event camera,'' in \emph{Computer Vision--ECCV 2016: 14th European Conference, Amsterdam, The Netherlands, October 11-14, 2016, Proceedings, Part VI 14}.\hskip 1em plus 0.5em minus 0.4em\relax Springer, 2016, pp. 349--364.

\bibitem{vidal2018ultimate}
A.~R. Vidal, H.~Rebecq, T.~Horstschaefer, and D.~Scaramuzza, ``Ultimate slam? combining events, images, and imu for robust visual slam in hdr and high-speed scenarios,'' \emph{IEEE Robotics and Automation Letters}, vol.~3, no.~2, pp. 994--1001, 2018.

\bibitem{pan2019bringing}
L.~Pan, C.~Scheerlinck, X.~Yu, R.~Hartley, M.~Liu, and Y.~Dai, ``Bringing a blurry frame alive at high frame-rate with an event camera,'' in \emph{Proceedings of the IEEE/CVF Conference on Computer Vision and Pattern Recognition}, 2019, pp. 6820--6829.

\bibitem{chin2019star}
T.-J. Chin, S.~Bagchi, A.~Eriksson, and A.~Van~Schaik, ``Star tracking using an event camera,'' in \emph{Proceedings of the IEEE/CVF Conference on Computer Vision and Pattern Recognition Workshops}, 2019, pp. 0--0.

\bibitem{orchard2015converting}
G.~Orchard, A.~Jayawant, G.~K. Cohen, and N.~Thakor, ``Converting static image datasets to spiking neuromorphic datasets using saccades,'' \emph{Frontiers in neuroscience}, vol.~9, p. 437, 2015.

\bibitem{li2017cifar10}
H.~Li, H.~Liu, X.~Ji, G.~Li, and L.~Shi, ``Cifar10-dvs: an event-stream dataset for object classification,'' \emph{Frontiers in neuroscience}, vol.~11, p. 309, 2017.

\bibitem{amir2017low}
A.~Amir, B.~Taba, D.~Berg, T.~Melano, J.~McKinstry, C.~Di~Nolfo, T.~Nayak, A.~Andreopoulos, G.~Garreau, M.~Mendoza \emph{et~al.}, ``A low power, fully event-based gesture recognition system,'' in \emph{Proceedings of the IEEE conference on computer vision and pattern recognition}, 2017, pp. 7243--7252.

\bibitem{wang2019ev}
Y.~Wang, B.~Du, Y.~Shen, K.~Wu, G.~Zhao, J.~Sun, and H.~Wen, ``Ev-gait: Event-based robust gait recognition using dynamic vision sensors,'' in \emph{Proceedings of the IEEE/CVF Conference on Computer Vision and Pattern Recognition}, 2019, pp. 6358--6367.

\bibitem{jin2018hybrid}
Y.~Jin, W.~Zhang, and P.~Li, ``Hybrid macro/micro level backpropagation for training deep spiking neural networks,'' \emph{Advances in neural information processing systems}, vol.~31, 2018.

\bibitem{shrestha2018slayer}
S.~B. Shrestha and G.~Orchard, ``Slayer: Spike layer error reassignment in time,'' \emph{Advances in neural information processing systems}, vol.~31, 2018.

\bibitem{wu2019direct}
Y.~Wu, L.~Deng, G.~Li, J.~Zhu, Y.~Xie, and L.~Shi, ``Direct training for spiking neural networks: Faster, larger, better,'' in \emph{Proceedings of the AAAI conference on artificial intelligence}, vol.~33, no.~01, 2019, pp. 1311--1318.

\bibitem{deng2020rethinking}
L.~Deng, Y.~Wu, X.~Hu, L.~Liang, Y.~Ding, G.~Li, G.~Zhao, P.~Li, and Y.~Xie, ``Rethinking the performance comparison between snns and anns,'' \emph{Neural networks}, vol. 121, pp. 294--307, 2020.

\bibitem{wu2021tandem}
J.~Wu, Y.~Chua, M.~Zhang, G.~Li, H.~Li, and K.~C. Tan, ``A tandem learning rule for effective training and rapid inference of deep spiking neural networks,'' \emph{IEEE Transactions on Neural Networks and Learning Systems}, 2021.

\bibitem{qiao2023batch}
G.~Qiao, N.~Ning, Y.~Zuo, P.~Zhou, M.~Sun, S.~Hu, Q.~Yu, and Y.~Liu, ``Batch normalization-free weight-binarized snn based on hardware-saving if neuron,'' \emph{Neurocomputing}, p. 126234, 2023.

\bibitem{kugele2020efficient}
A.~Kugele, T.~Pfeil, M.~Pfeiffer, and E.~Chicca, ``Efficient processing of spatio-temporal data streams with spiking neural networks,'' \emph{Frontiers in Neuroscience}, vol.~14, p. 439, 2020.

\bibitem{kim2021revisiting}
Y.~Kim and P.~Panda, ``Revisiting batch normalization for training low-latency deep spiking neural networks from scratch,'' \emph{Frontiers in neuroscience}, p. 1638, 2021.

\bibitem{fang2021deep}
W.~Fang, Z.~Yu, Y.~Chen, T.~Huang, T.~Masquelier, and Y.~Tian, ``Deep residual learning in spiking neural networks,'' \emph{Advances in Neural Information Processing Systems}, vol.~34, pp. 21\,056--21\,069, 2021.

\bibitem{wu2021liaf}
Z.~Wu, H.~Zhang, Y.~Lin, G.~Li, M.~Wang, and Y.~Tang, ``Liaf-net: Leaky integrate and analog fire network for lightweight and efficient spatiotemporal information processing,'' \emph{IEEE Transactions on Neural Networks and Learning Systems}, vol.~33, no.~11, pp. 6249--6262, 2021.

\bibitem{yao2021temporal}
M.~Yao, H.~Gao, G.~Zhao, D.~Wang, Y.~Lin, Z.~Yang, and G.~Li, ``Temporal-wise attention spiking neural networks for event streams classification,'' in \emph{Proceedings of the IEEE/CVF International Conference on Computer Vision}, 2021, pp. 10\,221--10\,230.

\bibitem{wang2021event}
Y.~Wang, X.~Zhang, Y.~Shen, B.~Du, G.~Zhao, L.~Cui, and H.~Wen, ``Event-stream representation for human gaits identification using deep neural networks,'' \emph{IEEE Transactions on Pattern Analysis and Machine Intelligence}, vol.~44, no.~7, pp. 3436--3449, 2021.

\bibitem{fang2021incorporating}
W.~Fang, Z.~Yu, Y.~Chen, T.~Masquelier, T.~Huang, and Y.~Tian, ``Incorporating learnable membrane time constant to enhance learning of spiking neural networks,'' in \emph{Proceedings of the IEEE/CVF International Conference on Computer Vision}, 2021, pp. 2661--2671.

\bibitem{lopez2013efectividad}
M.~M. L{\'o}pez-Rodr{\'\i}guez, M.~Fern{\'a}ndez-Mart{\'\i}nez, G.~A. Matar{\'a}n-Pe{\~n}arrocha, M.~E. Rodr{\'\i}guez-Ferrer, G.~G. G{\'a}mez, and E.~A. Ferr{\'a}ndiz, ``Efectividad de la biodanza acu{\'a}tica sobre la calidad del sue{\~n}o, la ansiedad y otros s{\'\i}ntomas en pacientes con fibromialgia,'' \emph{Medicina Cl{\'\i}nica}, vol. 141, no.~11, pp. 471--478, 2013.

\bibitem{cheng2017effects}
S.-L. Cheng, H.-F. Sun, and M.-L. Yeh, ``Effects of an 8-week aerobic dance program on health-related fitness in patients with schizophrenia,'' \emph{journal of nursing research}, vol.~25, no.~6, pp. 429--435, 2017.

\bibitem{marquez2017regular}
D.~X. Marquez, R.~Wilson, S.~Agui{\~n}aga, P.~V{\'a}squez, L.~Fogg, Z.~Yang, J.~Wilbur, S.~Hughes, and C.~Spanbauer, ``Regular latin dancing and health education may improve cognition of late middle-aged and older latinos,'' \emph{Journal of aging and physical activity}, vol.~25, no.~3, pp. 482--489, 2017.

\bibitem{Chen2022AnatomyAware3H}
T.~Chen, C.~Fang, X.~Shen, Y.~Zhu, Z.~Chen, and J.~Luo, ``Anatomy-aware 3d human pose estimation with bone-based pose decomposition,'' \emph{IEEE Transactions on Circuits and Systems for Video Technology}, vol.~32, pp. 198--209, 2022.

\bibitem{Chen2020MonocularHP}
Y.~Chen, Y.~Tian, and M.~He, ``Monocular human pose estimation: A survey of deep learning-based methods,'' \emph{Comput. Vis. Image Underst.}, vol. 192, p. 102897, 2020.

\bibitem{Hassan2019Resolving3H}
M.~Hassan, V.~Choutas, D.~Tzionas, and M.~J. Black, ``Resolving 3d human pose ambiguities with 3d scene constraints,'' \emph{2019 IEEE/CVF International Conference on Computer Vision (ICCV)}, pp. 2282--2292, 2019.

\bibitem{zhang2023neuromorphic}
Z.~Zhang, K.~Chai, H.~Yu, R.~Majaj, F.~Walsh, E.~Wang, U.~Mahbub, H.~Siegelmann, D.~Kim, and T.~Rahman, ``Neuromorphic high-frequency 3d dancing pose estimation in dynamic environment,'' \emph{Neurocomputing}, p. 126388, 2023.

\end{thebibliography}

\end{document}